\documentclass[10pt,twocolumn,letterpaper]{article}

\usepackage[pagenumbers]{cvpr} 

\usepackage[dvipsnames]{xcolor}

\usepackage{graphicx}
\usepackage{amsmath}
\usepackage{amssymb}
\usepackage{booktabs}
\usepackage{makecell, multirow, tabularx}
\usepackage{enumitem} 

\definecolor{cvprblue}{rgb}{0.21,0.49,0.74}
\usepackage[pagebackref,breaklinks,colorlinks,citecolor=cvprblue]{hyperref}

\def\paperID{17636} 
\def\confName{CVPR}
\def\confYear{2024}

\title{A Contrastive Compositional Benchmark for Text-to-Image Synthesis: A Study with Unified Text-to-Image Fidelity Metrics}

\author{%
Xiangru Zhu\textsuperscript{1,3}\thanks{Work done during the internship at Alibaba.}, 
        Penglei Sun\textsuperscript{2},
        Chengyu Wang\textsuperscript{3},
        Jingping Liu\textsuperscript{4},
        Zhixu Li\textsuperscript{1}\thanks{Corresponding author},
        Yanghua Xiao\textsuperscript{1}\footnotemark[2],
        Jun Huang\textsuperscript{3}
        \\
        \textsuperscript{1}Fudan University, 
        \textsuperscript{2}The Hong Kong University of Science and Technology (Guangzhou), \\
        \textsuperscript{3}Alibaba Group,
        \textsuperscript{4}East China University of Science and Technology
        \\
        {\tt\small \{xrzhu19, zhixuli, shawyh\}@fudan.edu.cn, psun012@connect.hkust-gz.edu.cn} \\
        { \tt \small jingpingliu@ecust.edu.cn, \{chengyu.wcy,huangjun.hj\}@alibaba-inc.com}
}

\begin{document}
\maketitle
\begin{abstract}
Text-to-image (T2I) synthesis has recently achieved significant advancements. However, challenges remain in the model's compositionality, which is the ability to create new combinations from known components. 
We introduce Winoground-T2I, a benchmark designed to evaluate the compositionality of T2I models. This benchmark includes 11K complex, high-quality contrastive sentence pairs spanning 20 categories. These contrastive sentence pairs with subtle differences enable fine-grained evaluations of T2I synthesis models. 
Additionally, to address the inconsistency across different metrics, we propose a strategy that evaluates the reliability of various metrics by using comparative sentence pairs.
%
%
We use Winoground-T2I with a dual objective: to evaluate the performance of T2I models and the metrics used for their evaluation. Finally, we provide insights into the strengths and weaknesses of these metrics and the capabilities of current T2I models in tackling challenges across a range of complex compositional categories.
Our benchmark is publicly available at \href{https://github.com/zhuxiangru/Winoground-T2I}{https://github.com/zhuxiangru/Winoground-T2I}.
\end{abstract}
\section{Introduction}
\label{sec:intro}

\begin{figure}[ht]
  \centering
    \begin{subfigure}{.25\textwidth}
        \centering
        \includegraphics[width=1\linewidth]{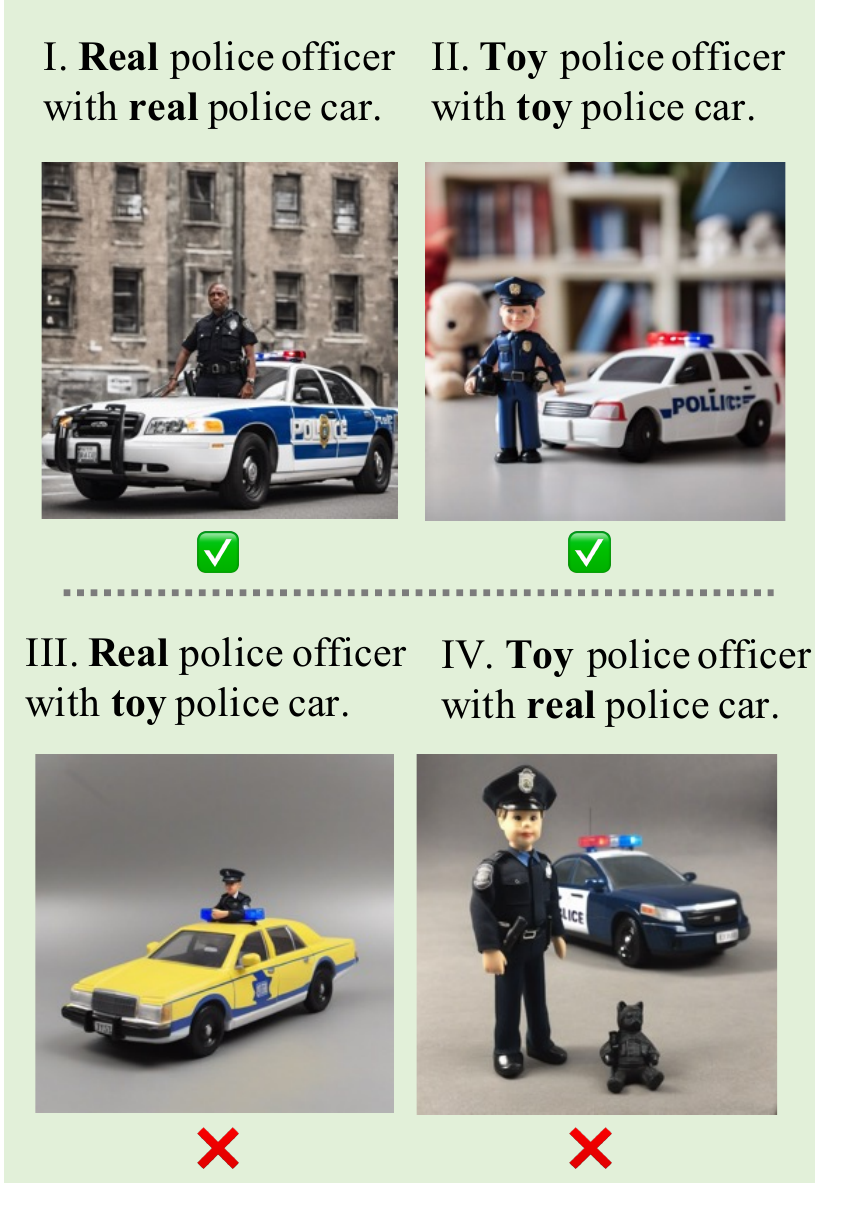}
        \caption{ }
        \label{fig:failure_case_sdxl_a}
    \end{subfigure}%
    \begin{subfigure}{.25\textwidth}
        \centering
        \includegraphics[width=1\linewidth]{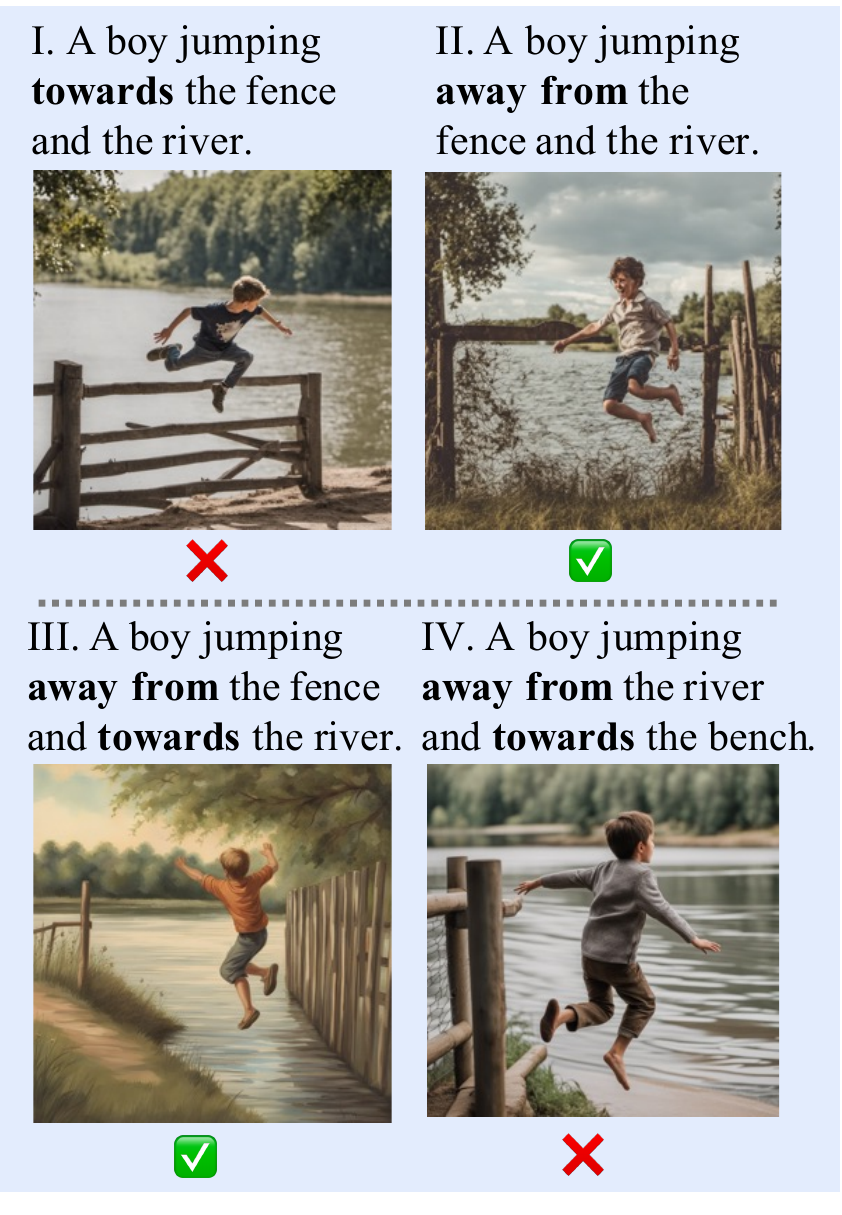}
        \caption{ }
        \label{fig:failure_case_sdxl_b}
    \end{subfigure}
  \caption{
  Failed cases on Stable Diffusion XL 1.0. I and II in (a) and (b) are about a single attribute or direction, while III and IV are novel combinations of known components. In (a), the model has the ability to render \emph{real} and \emph{toy} accurately in I and II, but fails in III and IV. In (b), the model can only render \emph{away from the fence} and fails in rendering \emph{towards the fence}, as shown in the images of I and II. The model always tends to generate \emph{away from the fence} whatever the combinations are, as shown in III and IV.
  }
  \label{fig:example}
\end{figure} 

Text-to-image (T2I) synthesis, which generates images from text prompts, has advanced significantly in art, gaming, and design. Models like Stable Diffusion~\cite{DBLP:conf/cvpr/RombachBLEO22, DBLP:journals/corr/abs-2307-01952}, Midjourney and DALL-E 3~\cite{dalle3} have seen extensive application across various scenarios.

However, current T2I synthesis models struggle to mirror compositional text prompts accurately~\cite{DBLP:journals/corr/abs-2305-13921, DBLP:journals/tog/CheferAVWC23, DBLP:conf/iclr/FengHFJANBWW23}, which involves understanding and generating novel combinations from known components~\cite{chomsky2014aspects, DBLP:conf/iclr/KeysersSSBFKMSS20, DBLP:conf/eccv/LiuLDTT22}. 
For example, in Figure~\ref{fig:failure_case_sdxl_b}, the Stable Diffusion XL 1.0 successfully renders \emph{real} and \emph{toy} police officers and cars in separate instances, but fails in their combinations. Similarly, in Figure~\ref{fig:failure_case_sdxl_b}, the model always renders \emph{away from the fence} regardless of the combination variations, indicating a tendency to favor familiar combinations.
%
%
This limitation is thought to derive from inadequate text representation~\cite{DBLP:conf/eccv/LiuLDTT22} and the lack of benchmarks that effectively measure the learning of compositionality in realistic scenarios~\cite{DBLP:conf/iclr/KeysersSSBFKMSS20}.
%
%

\begin{table*}[ht]
\scriptsize
\centering
\setlength\tabcolsep{3pt} 
\begin{tabular}{cccccccc}
\specialrule{.1em}{.05em}{.05em} 
\textbf{Benchmark} & \textbf{\begin{tabular}[c]{@{}l@{}}Source\end{tabular}} & \textbf{\begin{tabular}[c]{@{}l@{}}Comp.\end{tabular}} & \textbf{\begin{tabular}[c]{@{}l@{}}Quality\\Control\end{tabular}} &  \textbf{\begin{tabular}[c]{@{}l@{}}Contr.\end{tabular}} & \textbf{\begin{tabular}[c]{@{}l@{}}Compositional Categories\end{tabular}} & \textbf{\begin{tabular}[c]{@{}l@{}}Sample\\Count\end{tabular}} & \textbf{\begin{tabular}[c]{@{}l@{}}Comp.\\Category\\(All)\end{tabular}} \\
\hline
DrawBench~\cite{DBLP:conf/nips/SahariaCSLWDGLA22} & human & $\times$ & $\checkmark$ & $\times$ & \begin{tabular}[c]{@{}c@{}}color, spatial, action\end{tabular} & 200 & 3(11) \\
\hline
PartiPromps~\cite{DBLP:journals/tmlr/YuXKLBWVKYAHHPLZBW22} & human & $\times$ & $\checkmark$ & $\times$ & \begin{tabular}[c]{@{}c@{}}property\&position, quantity\end{tabular} & 1600 & 2(11)\\
\hline
\begin{tabular}[c]{@{}c@{}}C-CUB/Flowers\\ Color/Shape\cite{DBLP:conf/nips/ParkALDR21}\end{tabular} & COCO~\cite{DBLP:conf/eccv/LinMBHPRDZ14} & $\checkmark$ & -- & $\times$ & color, shape & 346.8K & 2 \\
\hline
SR\_2D~\cite{DBLP:conf/iclr/FengHFJANBWW23} & COCO~\cite{DBLP:conf/eccv/LinMBHPRDZ14} & $\checkmark$ & -- & $\times$ & \begin{tabular}[c]{@{}c@{}}spatial\end{tabular} & 25,280 & 1 \\ 
\hline
PaintSkills\cite{DBLP:journals/corr/abs-2202-04053} & template & $\checkmark$ & -- & $\times$ & \begin{tabular}[c]{@{}c@{}}object,counting,spatial\end{tabular} & \begin{tabular}[c]{@{}c@{}}65535\end{tabular} & 2(3)\\ 
\hline
ABC-6K\cite{DBLP:conf/iclr/FengHFJANBWW23} & COCO~\cite{DBLP:conf/eccv/LinMBHPRDZ14} & $\checkmark$ & -- & $\times$ & color & 6.4K & 1 \\
\hline
CC-500\cite{DBLP:conf/iclr/FengHFJANBWW23} & template & $\checkmark$ & -- & $\times$ & color & 500 & 1 \\
\hline
TIFA v1.0\cite{DBLP:journals/corr/abs-2303-11897} & \begin{tabular}[c]{@{}c@{}}COCO~\cite{DBLP:conf/eccv/LinMBHPRDZ14},DrawBench~\cite{DBLP:conf/nips/SahariaCSLWDGLA22},\\ PartiPrompts~\cite{DBLP:journals/tmlr/YuXKLBWVKYAHHPLZBW22},PaintSkill\cite{DBLP:journals/corr/abs-2202-04053}\end{tabular} & $\checkmark$ & -- & $\times$ & \begin{tabular}[c]{@{}c@{}}attribute, spatial, action, location, color, counting, material, shape, etc.\end{tabular} & 4081 & 8(12) \\ 
\hline
VPEval-skill\cite{DBLP:journals/corr/abs-2305-15328} & COCO~\cite{DBLP:conf/eccv/LinMBHPRDZ14} & $\checkmark$ & -- & $\times$ & \begin{tabular}[c]{@{}c@{}}object, spatial, count, scale, text rendering\end{tabular} & \begin{tabular}[c]{@{}c@{}}3803\end{tabular} & 3(5) \\ 
\hline
DSG-1K~\cite{DBLP:journals/corr/abs-2310-18235} & COCO~\cite{DBLP:conf/eccv/LinMBHPRDZ14} & $\checkmark$ & -- & $\times$ & \begin{tabular}[c]{@{}c@{}}color, material, count, texture, shape, spatial, action, etc\end{tabular} & 1060 & 8(13)\\
\hline
\begin{tabular}[c]{@{}c@{}}HRS-Bench\\(composition part)~\cite{DBLP:journals/corr/abs-2304-05390}\end{tabular} & template \& LLM & $\checkmark$ & $\times$ & $\times$ & \begin{tabular}[c]{@{}c@{}}spacial, size, color, action, counting\end{tabular} & 6,000 & 5\\
\hline
\begin{tabular}[c]{@{}c@{}}T2I-CompBench\cite{DBLP:journals/corr/abs-2307-06350}\end{tabular} & LLM & $\checkmark$ & $\times$ & $\times$ & \begin{tabular}[c]{@{}c@{}}color, shape, texture,spatial, action, complex\end{tabular} & 6,000 & 6 \\
\hline
Winoground-T2I & LLM & $\checkmark$ & $\checkmark$ & $\checkmark$ & \begin{tabular}[c]{@{}c@{}}action, spatial, direction, color, number, size, texture, shape, age,\\weight, manner, sentiment, procedure, speed, etc. \end{tabular} & 22K & 20 \\
\specialrule{.1em}{.05em}{.05em} 
\end{tabular}
\caption{
The comparison of compositional T2I benchmarks. \emph{Comp.} is the abbreviation of \emph{Compositional}. \emph{Contr.} is the abbreviation of \emph{Contrastive}. The categories in PromptParts are chosen based on the dataset's Aspects rather than Categories.
\label{tab:benchmarks}
}
\end{table*}

A benchmark that realistically reflects diverse combinations of known components is crucial.
Current benchmarks, however, have several shortcomings. 
First, all existing benchmarks \textbf{lack fine-grained samples for testing subtle differences}
; yet, those benchmarks that rely on more hackable modifications, such as word replacement or addition, may not effectively measure a model's true capability in understanding the cross-modal compositionality~\cite{DBLP:conf/iclr/Yuksekgonul0KJ023,DBLP:journals/corr/abs-2306-14610}.
Second, there is \textbf{a gap between the complexity of benchmark and real-world}. 
In T2I-CompBench~\cite{DBLP:journals/corr/abs-2307-06350}, 5 out of 6 cases are single-category, with only 1 being multi-category.
In PaintSkills~\cite{DBLP:journals/corr/abs-2202-04053}, VPEval-skill\cite{DBLP:journals/corr/abs-2305-15328} and DSG~\cite{DBLP:journals/corr/abs-2310-18235}, non-compositional categories like \emph{object}, \emph{text rendering} and \emph{global information} comprise 40\%, 21\%, and 52\% of cases, respectively. 
Third, some benchmarks \textbf{lack prompt quality control}. 
In 100 cases randomly sampled from \emph{complex} category of T2I-CompBench~\cite{DBLP:journals/corr/abs-2307-06350}, 22\% of prompts are subjective or contradictory, 
such as the prompt \emph{The jagged zigzag was nestled between the smooth curve and the curved line}.
Last, many benchmarks have \textbf{a narrow range of categories}. Early ones have only 1-2 categories~\cite{DBLP:conf/nips/ParkALDR21, DBLP:conf/iclr/FengHFJANBWW23, DBLP:journals/corr/abs-2212-10015, DBLP:journals/corr/abs-2212-10015, DBLP:journals/corr/abs-2202-04053}, while current ones like DSG-1K~\cite{DBLP:journals/corr/abs-2310-18235} include up to 13 categories; however, 4 of these are non-compositional, a limitation also observed in TIFA~\cite{DBLP:journals/corr/abs-2303-11897} and VPEval-skill~\cite{DBLP:journals/corr/abs-2305-15328}.

We introduce Winoground-T2I, a benchmark with contrastive sentence pairs, to evaluate the compositionality of T2I synthesis. 
%
Its advantages include:
1) \textbf{Subtle differences within contrastive sample pairs}. Our benchmark consists of contrastive sentence pairs, where the elements of two sentences are the same, and the only difference within a pair is the order of words. 
2) \textbf{Samples closer to the real world}. Most sentence pairs span multiple compositional categories, attempting to intimate the complex prompts in real scenarios.
3) \textbf{High quality}. We establish rigorous criteria to identify and exclude invalid LLM-generated contrastive sample pairs, filtering them with automated and human review.
4) \textbf{Wide range of categories}. Our benchmark covers 20 categories with a total of 22K sentences.

%
%

Another critical issue with T2I synthesis evaluation concerns the metrics used. We note that these metrics often produce inconsistent and conflicting results, highlighting the need for a reliable evaluation method.
We propose a strategy to evaluate the reliability of these metrics from 4 perspectives: inter-pair alignment with human preferences, intra-pair consistency, and discrimination, stability, and efficiency. 
With our benchmark and a reliable metric we selected, we analyze the strengths and weaknesses of current popular T2I models across various categories.


Our contributions are summarised as follows:
\begin{itemize}[itemsep=0pt] 
    \item We introduce Winoground-T2I, a benchmark for evaluating the compositionality of T2I synthesis models, including 11K diverse, complex, high-quality, contrastive sentence pairs across 20 categories.

    \item We propose a strategy for evaluating the reliability of T2I evaluation metrics, focusing on not only traditional inter-pair comparison, but also intra-pair comparison, stability, and efficiency. 
    
    \item We analyze current mainstream T2I synthesis models, highlighting strengths in generating attributes such as \emph{color}, \emph{material}, and \emph{spatial}, and identifying room for improvement in less common attributes and relations which are overlooked in previous benchmarks.

\end{itemize}





\section{Related Work}
\label{sec:related_work}

\noindent 
\textbf{Compositionality Benchmarks for T2I synthesis}. Early benchmarks like DrawBench~\cite{DBLP:conf/nips/SahariaCSLWDGLA22} and PartiPromps~\cite{DBLP:journals/tmlr/YuXKLBWVKYAHHPLZBW22} focus on general capabilities rather than compositionality and are limited in scale and human annotations. Later benchmarks, such as C-CUB/Flowers Color/Shape~\cite{DBLP:conf/nips/ParkALDR21}, SR\_2D~\cite{DBLP:conf/iclr/FengHFJANBWW23}, ABC-6K~\cite{DBLP:conf/nips/ParkALDR21}, TIFA v1.0~\cite{DBLP:journals/corr/abs-2303-11897}, and VPEval-skill~\cite{DBLP:journals/corr/abs-2305-15328}, focus on compositionality and utilize specific attributes or relationships from datasets like MSCOCO~\cite{DBLP:conf/eccv/LinMBHPRDZ14}. 
However, these benchmarks need to be improved in diversity due to the constraints of the existing sample pool. 
Efforts like PaintSkills~\cite{DBLP:journals/corr/abs-2202-04053} and CC-500~\cite{DBLP:conf/iclr/FengHFJANBWW23} try to improve diversity by filling the slot in templates with dataset vocabulary but are still restricted by the available words. HRS-Bench(compositional part)~\cite{DBLP:journals/corr/abs-2304-05390} and T2I-CompBench~\cite{DBLP:journals/corr/abs-2307-06350} overcome vocabulary constraints by using LLMs to fill the slots. 
However, they introduce a new issue: it's difficult to identify either unreasonable or non-visual sentences because the standards of reasonable and visual in T2I synthesis differ from those in the real world. 

\noindent 
\textbf{Evaluation Metrics for T2I synthesis}. 
Traditional metrics face challenges in evaluating text-image fidelity in T2I synthesis. Inception Score (IS)~\cite{DBLP:journals/corr/abs-1801-01973} and Frechet Inception Distance (FID)~\cite{DBLP:conf/nips/HeuselRUNH17} are incapable of zero-shot evaluation. Metrics like CLIPScore~\cite{DBLP:conf/emnlp/HesselHFBC21} and BLIP-ITM~\cite{DBLP:conf/icml/0001LXH22, DBLP:conf/icml/0008LSH23}, focusing on global visual-textual similarity, along with BLIP-ITC~\cite{DBLP:conf/icml/0001LXH22, DBLP:conf/icml/0008LSH23}, METEOR~\cite{DBLP:conf/acl/BanerjeeL05} and AC-T2I~\cite{DBLP:journals/corr/abs-2304-05390} (relying on CIDEr~\cite{DBLP:conf/cvpr/VedantamZP15} and BLUE~\cite{DBLP:conf/acl/PapineniRWZ02}), focusing on global textual similarity, provide coarse-grained scores, falling short in fine-grained T2I fidelity evaluation. 


T2I fidelity evaluation research can be divided into three categories: 1) \emph{Feature Based Metrics}, aligning with human preference by through reinforcement learning from human feedback (RLHF) building on CLIP~\cite{DBLP:conf/icml/RadfordKHRGASAM21} or BLIP~\cite{DBLP:conf/icml/0001LXH22}, such as PickScore~\cite{DBLP:journals/corr/abs-2305-01569} and ImageReward~\cite{DBLP:journals/corr/abs-2304-05977}. 2) \emph{Visual Programming Based Metrics}, decomposing prompts into questions for evaluation, including VQA-based ones (e.g., TIFA~\cite{DBLP:journals/corr/abs-2303-11897}, DSG~\cite{DBLP:journals/corr/abs-2310-18235}) and tool-based ones (e.g., VPEval~\cite{DBLP:journals/corr/abs-2305-15328}). 3) \emph{LLM's Chain of Thought (CoT) Based Metrics}, using LLMs's CoT for scoring and rationale prediction, such as LLMScore~\cite{DBLP:journals/corr/abs-2305-11116} and MiniGPT4-CoT~\cite{DBLP:journals/corr/abs-2307-06350}.


\section{Problem Formulation}\label{sec:Problem Formulation}

\noindent
\textbf{Input}. 
A textual prompt $t$ can be decomposed into multiple semantic fragments $t=\{s_1, \ldots, s_n\}$, each contributing to its overall meaning. 
An image $v$ is generated by a T2I synthesis model with the input $t$. 
%

\noindent
\textbf{Output}. 
A fidelity score $y(t, v)$ measures how well image $v$ represents the collective semantics of prompt $t$. 
The evaluation process consists of two parts. 
First, it measures the distance $d(s_j, v)$ for each semantic fragment $s_j$ in the prompt $t$. The distance varies according to compositional categories like actions, locations, and attributes. 
Second, it aggregates these distances with a scoring function $g(\cdot)$ to compute the overall score for the input $t$ as $y(t,v)=g(\{d(s_j, v)\}_{j=1}^n)$, such as maximum, average, weighted sum or neural networks.


%

\section{Dataset}

\begin{figure}[t]
  \centering
  \includegraphics[width=1\linewidth]{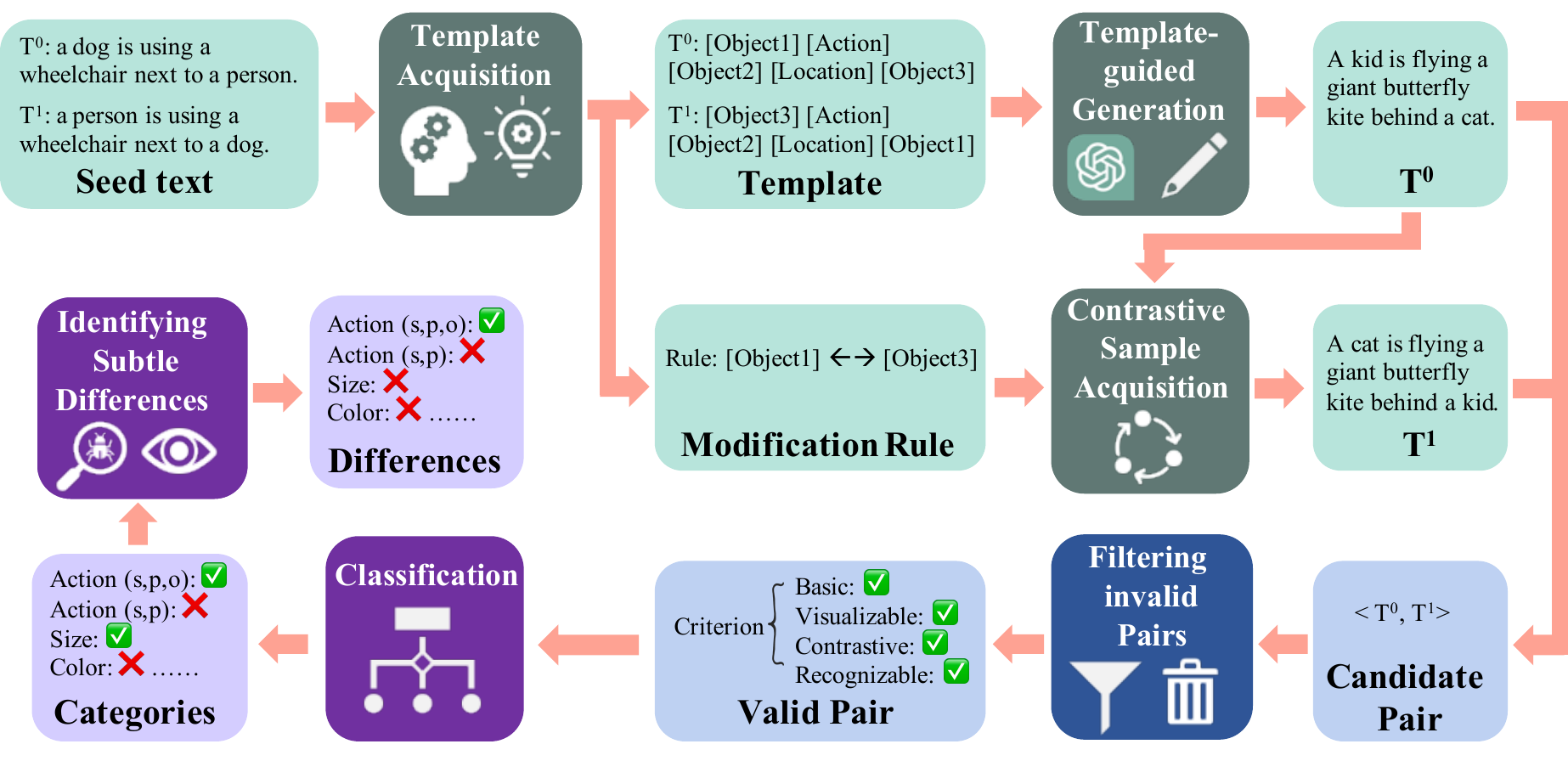}
  \caption{The procedure of data collection, quality control and labeling. The light represents the data format, and the dark represents the modules. The green is data collection, the blue is quality control, the purple is labeling, and the red is data flow.  
  }
  \label{fig:collection}
\end{figure} 

We utilize the pipeline in Figure~\ref{fig:collection} to automatically generate high-quality compositional contrastive sentence pairs, including \emph{Data Collection}, \emph{Quality Control} and \emph{Labeling}.


\subsection{Definition}\label{sec:definition}

\noindent
\textbf{Contrastive Samples}. 
Contrastive sentences in each pair $(T^0, T^1)$ should adhere to the following properties:
1) \textbf{Literal Similarity}: $T^0$ and $T^1$ are textually similar with the same elements (e.g., structures and words), differing in word order and negations to create distinct meanings. 
2) \textbf{Frequency Difference}: $T^1$ is no more common than $T^0$ in natural language usage.
3) \textbf{Reasonability}: $T^0$ and $T^1$ are semantically reasonable in real or fictional contexts, such as children's literature, animations, and science fiction.
4) \textbf{Visualizability}: $T^0$ and $T^1$ evoke vivid images.
5) \textbf{Discrimination}: The images evoked by $T^0$ and $T^1$ present distinguishable difference.


\subsection{Data Collection}\label{sec:Template-guided Generation}

\noindent
\textbf{Template Acquisition}. We choose 200 compositional sentence pairs labeled with \emph{no-tag}~\cite{DBLP:conf/emnlp/DiwanBCHM22} as seed pairs from Winoground~\cite{DBLP:conf/cvpr/ThrushJBSWKR22,DBLP:conf/emnlp/DiwanBCHM22}, which are typical compositional sentence pairs. These pairs are used to extract templates and modification rules manually. 
For example, considering the sentences \emph{a dog is using a wheelchair next to a person} and \emph{a person is using a wheelchair next to a dog}, the corresponding template and modification rule are:

{\small $T^0$ template: \emph{[Object1] [Action] [Object2] next to [Object3]}}

{\small $T^1$ template: \emph{[Object3] [Action] [Object2] next to [Object1]}}

{\small Modification Rule: \emph{[Object1]$\leftrightarrow$[Object3]}}





\noindent
\textbf{Template-guided Generation}. 
Utilizing the GPT-3.5, we generate diverse sentence pairs by filling the slots in two templates.
We discovered that while the sentence $T^0$ is typically diverse and accurate, the sentence $T^1$ often violates the modification rule. As a result, we restrict GPT-3.5 to only generating $T^0$ based on its prior knowledge and maximum likelihood. 
For example, guided by the above seed pair and the template of $T^0$, we generate a sentence $T^0$ that \emph{A kid is flying a giant butterfly kite behind a cat}.


\noindent
\textbf{Contrastive Sample Acquisition}. 
$T^1$ is derived by modifying the structure of $T^0$ following the modification rule, such as swapping or moving words. 
For example, we get a sentence as $T^1$ that \emph{a cat is flying a giant butterfly kite behind a kid} by swapping \emph{[Object1]} and \emph{[Object3]} in $T^0$.  
This method ensures that $T^1$ is a semantic variation of $T^0$, avoiding a random sequence or a semantically equivalent passive structure to $T^0$, which usually occurs in template-guided generation.
Notably, the $T^1$ is no more common due to its derivation from modification rule applied to $T^0$.


\subsection{Quality Control}\label{sec:Automatic Verification}

\emph{Literal similarity} and \emph{frequency difference} are assured by the 3-step collection. However, \emph{reasonability, visualizability} and \emph{discrimination} require additional consideration. 
%
For example, \emph{An engineer builds a bridge} and \emph{A bridge builds an engineer} are not both reasonable. As a result, filtering out invalid samples is crucial.
We define \emph{valid} texts for T2I models using 14 criteria as follows:

\noindent
\textbf{Basic}. Sentences should be 1) \emph{complete}, 2) \emph{with clear objects} and 3) \emph{reasonable semantic}. 

\noindent
\textbf{Visualizable}. Elements should be 4) \emph{image-depictable}, with global semantics representable in 5) \emph{static or multiple exposure scenes}.
Comparisons must be 6) \emph{visually quantifiable}, and maintain 7) \emph{a moderate level of detail}.
%

\noindent
\textbf{Contrastive}. Sentence pairs should 8) \emph{adhere to modification rules}, with 9) \emph{distinct textual semantics} and 10) \emph{visually distinguishable evoked images}.

\noindent
\textbf{Recognizable}. Both sentences should maintain key elements for the typical recognition of scenes and characters. This includes 11-14) \emph{item-specific, attire-specific, and action-specific scene and characters}.

\begin{figure}[t]
  \centering
  \begin{subfigure}[b]{\linewidth}
    \includegraphics[width=1\linewidth]
  {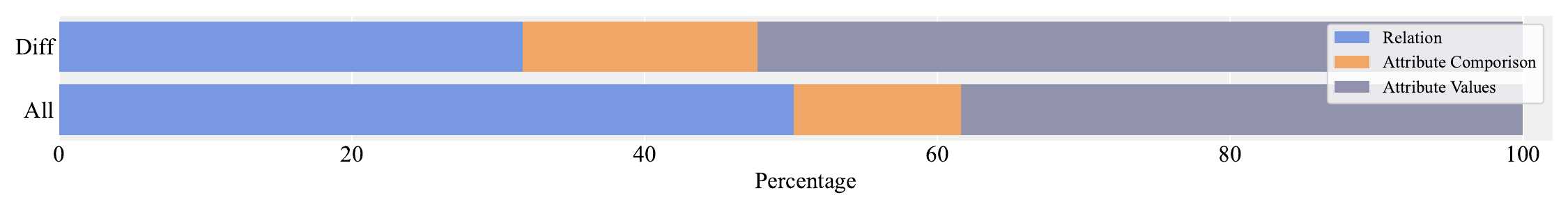}
    \caption{The distribution of 3 aspects of categories in \emph{All} and \emph{Diff} division.}
    \label{fig:category_percentage}
  \end{subfigure}
  \begin{subfigure}[b]{\linewidth}
    \includegraphics[width=1\linewidth]{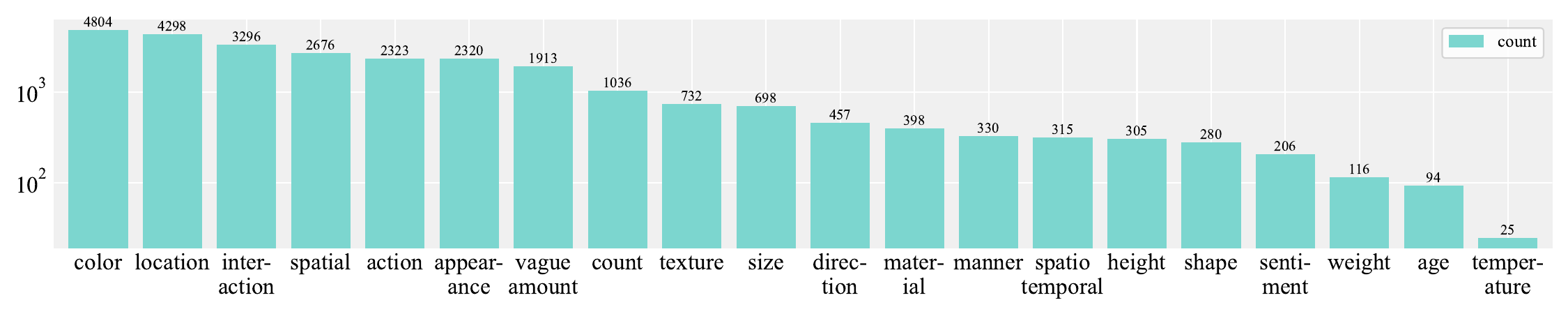}
    \caption{The distribution of all involved categories.}
    \label{fig:category_all}
  \end{subfigure}
  \begin{subfigure}[b]{\linewidth}
    \includegraphics[width=1\linewidth]{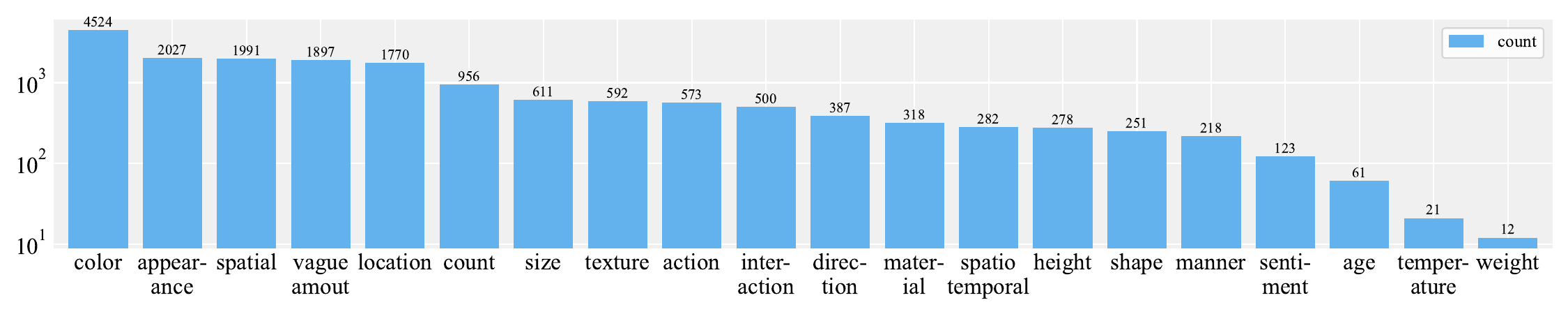}
    \caption{The distribution of difference within sample pairs.}
    \label{fig:category_diff}
  \end{subfigure}
  \caption{Statistics of categories.}
  \label{fig:category_distribution}
\end{figure}

\subsection{Labeling}
We employ two labeling methods, \textbf{All} and \textbf{Diff}, with shared labels to categorize our sentence pairs. These pairs are initially labeled by GPT-3.5 and then verified by humans. 
\emph{All} labels each pair with relevant categories, while \emph{Diff} labels each pair with categories that reveal contrastive elements between sentences. For each sentence pair, labels from \emph{Diff} are a subset of those from \emph{All}.
%
%
Due to most pairs falling into multiple categories, the total number of categorized pairs exceeds the actual count of pairs in our benchmark.

%

\subsection{Dataset Analysis}

\noindent
\textbf{Wide range of categories}\label{template_acquisition}. 
We categorize samples into 20 compositional categories, divided into three aspects: \emph{Relation} (e.g., \emph{Action, Interaction, Location, Spatial, Spatial-Temporal} and \emph{Direction}), \emph{Attribute Comparison} (e.g., \emph{Scale, Height, Weight} and \emph{Vague Amount}) and \emph{Attribute Values} (e.g., \emph{Counting, Color, Appearance, Texture, Material, Shape, Age, Sentiment, Temperature, Weight} and \emph{Manner}). 
The percentage of sentence pairs across 3 aspects is shown in Figure ~\ref{fig:category_percentage}, while the breakdown by individual categories is shown in Figure~\ref{fig:category_all}, where \emph{Action} refers to the action without following objects and \emph{Interaction} refers to the action with following objects.



\noindent
\textbf{Samples closer to the real world}. In Winoground-T2I, most sentences span multiple categories, which is closer to the prompts in realistic scenarios. Specifically, only 16\% of sentence pairs are tied to a single category. The proportions of pairs associated with 2, 3, 4, and 5 or more categories are respectively 47\%, 27\%, 8\% and 2\%. 
Though we label object-related sentences, we exclude \emph{Object} from our categories to avoid distribution imbalances and an overabundance of simplistic samples due to too many \emph{Object} category sentences.


\noindent
\textbf{High quality}. 
Following 14 guidelines for filtering, we utilized GPT-3.5 to filter these sentence pairs, eliminating over 42\% of the initial 48,344, resulting in 27,626 pairs. For quality assurance, 15 annotators and 2 checkers further manually reviewed the pairs, ultimately selecting 11,479 valid, non-duplicate contrastive samples.
We randomly sampled 100 cases from the filtered dataset and evaluated the quality of the 100 cases by 2 independent evaluators, different from the original annotators and checkers. We find that 99\% of the sentence pairs are valid. 
The reason for the invalid is the indistinguishable contrastive elements, colors \emph{fuchsia} and \emph{purple}.

\noindent
\textbf{Subtle differences within contrastive sample pairs}. 
With the labels from \emph{Diff}, 
56\% of the pairs contain only 1 challenging modification, 36\% for 2 modifications, and 8\% for more than 3 modifications. 
The percentage of sentence pairs across 3 aspects, labeled using the \emph{Diff} method, is shown in Figure ~\ref{fig:category_percentage}, while the distribution of categories of contrastive elements is shown in Figure ~\ref{fig:category_diff}. 
\section{Evaluating Metrics Reliability}\label{sec:reliable}

To comprehensively evaluate the reliability of various metrics, we develop statistical indicators from 4 different perspectives: \emph{Inter-pair Comparison}, \emph{Intra-pair Comparison}, \emph{Stability} and \emph{Efficiency}.

$t$ is a textual prompt and $v$ is its corresponding image generated by a T2I synthesis model. 
The automatic metric $y(t,v)$ and the human rating $h(t,v)$ both measure the fidelity of the generated $v$ for the input $t$. 
We measure the alignment of the metric $y$ with human preferences by the correlation. The correlation coefficient $r$ between metric score rankings and human rating rankings on a dataset of $m$ prompts is denoted as $r(\{y(t_i,v_i)\}_{i=1}^m, \{h(t_i,v_i)\}_{i=1}^m)$. 
For contrastive sentences $t^0$ and $t^1$ and their images $v^0$ and $v^1$, we denote the metric scores for $t^0$ as $Y^0 = \{y(t_i^0, v_i^0)\}_{i=1}^m$ and their corresponding human ratings as $H^0 = \{h(t_i^0, v_i^0)\}_{i=1}^m$. Similarly, we have $Y^1$ and $H^1$ for $t^1$. The correlation for $t^0$ and $t^1$ are computed separately.


\noindent
\textbf{Inter-pair Comparison} focuses on the automatic metric's alignment with human preference among different contrastive sentence pairs, a common approach in evaluating various T2I metrics in previous work. 
We measure the alignment between metric score rankings and human rating rankings among all $t^0$ is $r(Y^0, H^0)$ and that among all $t^1$ is $r(Y^1, H^1)$.

\noindent
\textbf{Intra-pair Comparison} focuses on the sensitivity to subtle differences within each contrastive sentence pair, which is reflected in two aspects: 

1) Two sentences in a pair ($t^0$, $t^1$), with closely similar human ratings, should receive similar rankings from the automatic metric.
%
Specifically, $Z=\{y(t_k,v_k) \mid k \in K\}$ is defined as the set of metric scores for image-text pairs that have closely similar human ratings, with $K=\{k \mid \arg_{k} h(t^0_k, v^0_k)=h(t^1_k, v^1_k)\}$ the index set for these pairs. 
Here, $Z^0$ and $Z^1$ represent the metric scores for the first and second sentences in these contrastive sentence pairs, respectively.
We represent the scores within a specified tolerance as similar automatic metric rankings.
The set of these score pairs from contrastive sentence pairs, which receive closely similar rankings, is denoted as 
%
$Z^{\prime}=\{y(t_k,v_k) \mid k \in K \cap K^{\prime}\}$, where the index set $K^{\prime}=\{k \mid \arg_{k} \lvert \frac{Y^0 - Y^1}{Y_{\max}-Y_{\min}} \lvert < \epsilon \}$ and $\epsilon$ is the tolerance. 
Finally, We measure 1) the correlation within these score pairs as $r_{same}=r(Z^0, Z^1)$, and 2) the proportion of similar rankings by the metric, denoted as $p_{same}=p(Z^{\prime} \mid Z)$.


2) Two sentences in a pair ($t^0$, $t^1$), with significant differences in human ratings, should also exhibit notable differences in the automatic metric rankings. 
%
This property reflects the metric's discriminability.
%
%
Specifically, we compare differences in metric scores between matching and mismatching sentence-image pairs. The set satisfing the discriminability is denoted as $\tilde{Z}=\{y(t_k, v_k) \mid k \in \tilde{K}\}$, where $\tilde{K}$ is the index set, $\tilde{K}=\{k \mid \arg_{k} y(t_k^0, v_k^0) > y(t_k^0, v_k^1) \text{ and } y(t_k^1, v_k^1) > y(t_k^1, v_k^0)\}$.
However, given the risk of misleading results caused by low-quality T2I synthesis, we prioritize high-quality evaluations by selecting sentence-image pairs with top metric scores. 
Specifically, we select those contrastive sentence pairs whose normalized scores exceed a threshold $\Omega$, emphasizing significant differences. The qualified set $Z^*=\{y(t_k, v_k) \mid k \in K \cap K^*\}$ comprising pairs satisfying the conditions, where the index set is $K^*=\{k \mid \arg_{k} \frac{Y^0 - Y_{\min}}{Y_{\max}-Y_{\min}} > \Omega \text{ and } \frac{Y^1 - Y_{\min}}{Y_{\max}-Y_{\min}} > \Omega\}$.
Finally, we measure the discriminability as the proportion of these discriminative pairs in the qualified set, denoted as $p_{diff}=p(\tilde{Z}|Z^*)$.


\noindent
\textbf{Stability} focuses on the reproducibility of results in repeated experiments. 
Specifically, we measure the stability by the average of correlation coefficients among repeated experiments. 
For example, 
$Y^{0(i)}$ and $Y^{1(i)}$ represent the scores from the i-th repetition for $Y^0$ and $Y^1$, respectively.
%
The average correlation coefficients for $Y^0$ is calculated as $\bar{r}=\frac{1}{T} \sum_{1 \leq i,j \leq T, i \neq j} r_{ij}(Y^{0(i)}, Y^{0(j)})$, where $T$ is the number of  repetitions. The same goes for $Y^1$.

\noindent
\textbf{Efficiency} focuses on the time cost for evaluating T2I fidelity, ensuring it remains within acceptable limits.
An oversimplified or insufficient evaluation may compromise a metric's reliability, often resulting from the excessive time cost.
%
%
We measure the efficiency by averaging time per sentence, denoted as $\frac{1}{m}\Sigma_{1}^m c(\{y(t_i,v_i)\})$, where $m$ is the number of sentences and $c(\cdot)$ represents the time taken.

\section{Experiment}

We attempt to answer four questions:

Q1: How challenging is our benchmark compared to other benchmarks? 

Q2: What is the performance of various open-source T2I synthesis models in our benchmark?

Q3: Which metric appears to be the most reliable, considering the discrepancies among diverse metrics? 

Q4: How do T2I models perform across various categories based on the optimal metric and our benchmark?

\subsection{Experiment Setting}

\noindent
\textbf{Baseline Benchmark} We compare our benchmark with current compositional ones: ABC-6K~\cite{DBLP:conf/iclr/FengHFJANBWW23}, CC-500~\cite{DBLP:conf/iclr/FengHFJANBWW23}, T2I-CompBench~\cite{DBLP:journals/corr/abs-2307-06350}, TIFA v1.0~\cite{DBLP:journals/corr/abs-2303-11897}, VPEval-skill~\cite{DBLP:journals/corr/abs-2305-15328}, DSG-1k~\cite{DBLP:journals/corr/abs-2310-18235}.  

\noindent
\textbf{T2I Synthesis Models} We generate images using Stable Diffusion v1.5~\cite{DBLP:conf/cvpr/RombachBLEO22} (denoted as SD1.5), Stable Diffusion v2.1~\cite{DBLP:conf/cvpr/RombachBLEO22} (denoted as SD2.1), Stable Diffusion XL v1.0~\cite{DBLP:journals/corr/abs-2307-01952} (denoted as SDXL) and DeepFloyd IF (an open source version of Imagen~\cite{DBLP:conf/nips/SahariaCSLWDGLA22}, denoted as IF). 
Specifically, SDXL and IF utilize stronger text encoders, compared to SD1.5 (CLIP ViT-L/14~\cite{DBLP:conf/icml/RadfordKHRGASAM21} with 123.65M parameters) and SD2.1 (CLIP ViT-H/14~\cite{DBLP:conf/icml/RadfordKHRGASAM21} with 354.0M parameters). 
SDXL uses the text encoders of CLIP ViT-bigG (694.7M parameters) together with CLIP ViT-L/14. 
IF uses the T5-XXL text encoder~\cite{DBLP:journals/jmlr/RaffelSRLNMZLL20} with 4.6B parameters. 


\noindent
\textbf{Evaluation Metrics}
We use following evaluation metrics. 
1) \emph{Feature Based Metrics}: CLIPScore~\cite{DBLP:conf/emnlp/HesselHFBC21}, BLIP-ITM Score~\cite{DBLP:conf/icml/0001LXH22, DBLP:conf/icml/0008LSH23}, PickScore~\cite{DBLP:journals/corr/abs-2305-01569}, ImageReward~\cite{DBLP:journals/corr/abs-2304-05977}. 
2) \emph{Visual Programming Based Metrics}: VPEval (utilizing external tools)~\cite{DBLP:journals/corr/abs-2305-15328}, TIFA~\cite{DBLP:journals/corr/abs-2303-11897}, DSG~\cite{DBLP:journals/corr/abs-2310-18235}. 
For the questions or tasks pre-generated by the LLM, we examine both fixed and unfixed ones to evaluate the stability and flexibility of these metric settings.
3) \emph{LLM's CoT Based Metrics}: MiniGPT-CoT~\cite{DBLP:journals/corr/abs-2307-06350} and LLMScore~\cite{DBLP:journals/corr/abs-2305-11116} and two improved versions of LLMScore, named as LLMScore$^+$ (using more detailed captions by a VLM) and LLMScore$^{++}$ (further reducing hallucinations by self-verification).


\noindent
\textbf{Human Rating}. The human rating on images generated by T2I models is used to evaluate the quality of automatic metrics. For correlation $r$ mentioned in Section \ref{sec:reliable}, we use Kendall's tau ($\tau$) and Spearman's rho ($\rho$), both ranging from -1 to 1.
A value closer to 1 indicates a positive correlation with human preferences, closer to -1 indicates a negative correlation, and closer to 0 suggests a weaker correlation.

For human rating, we create a subset called \emph{sample-100}, which consists of 100 sentence pairs $(T^0, T^1)$, randomly sampling 5-6 sentence pairs for each category. 
We generate one image per sentence from 200 sentences using 4 T2I synthesis models, resulting in 800 unique image-text pairs. 



\begin{figure}[ht]
  \centering
  
   \begin{subfigure}[b]{0.5\linewidth}
        \centering
        \includegraphics[width=\linewidth]{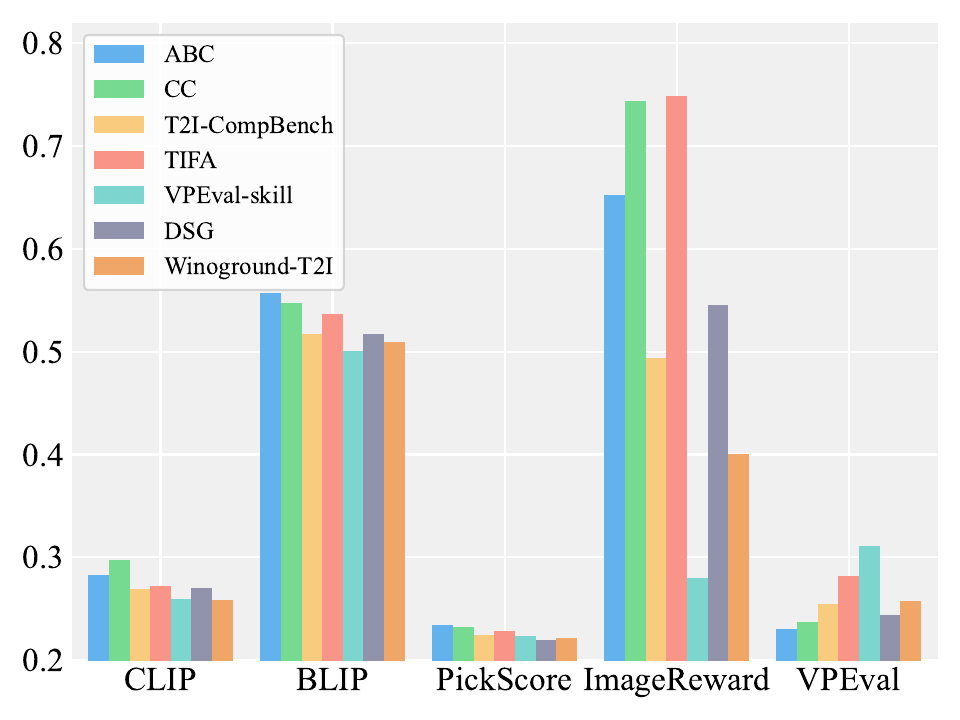}
        \label{fig:benchmark_result_1}
   \end{subfigure}%
   \begin{subfigure}[b]{0.5\linewidth}
        \centering
        \includegraphics[width=\linewidth]{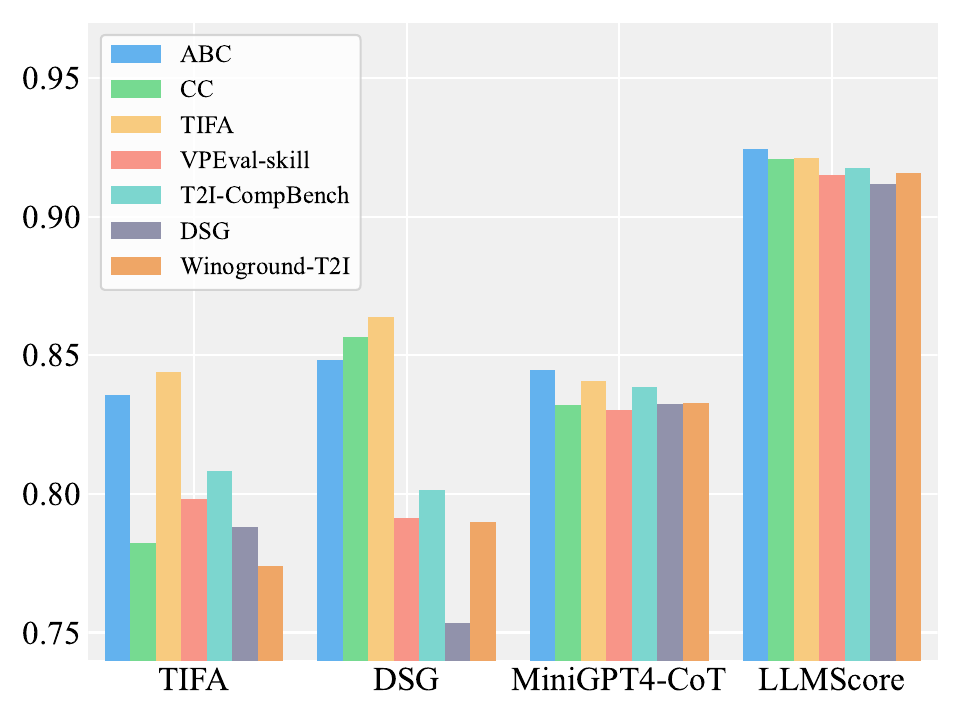}
        \label{fig:benchmark_result_2}
   \end{subfigure}
   \vspace{-10mm}
   \caption{Evaluation results from 7 benchmarks on SDXL.}
   \label{fig:benchmark_results}
\end{figure}

\begin{figure}[ht]
  \centering
  

    \begin{subfigure}[b]{0.5\linewidth}
        \centering
        \includegraphics[width=\linewidth]{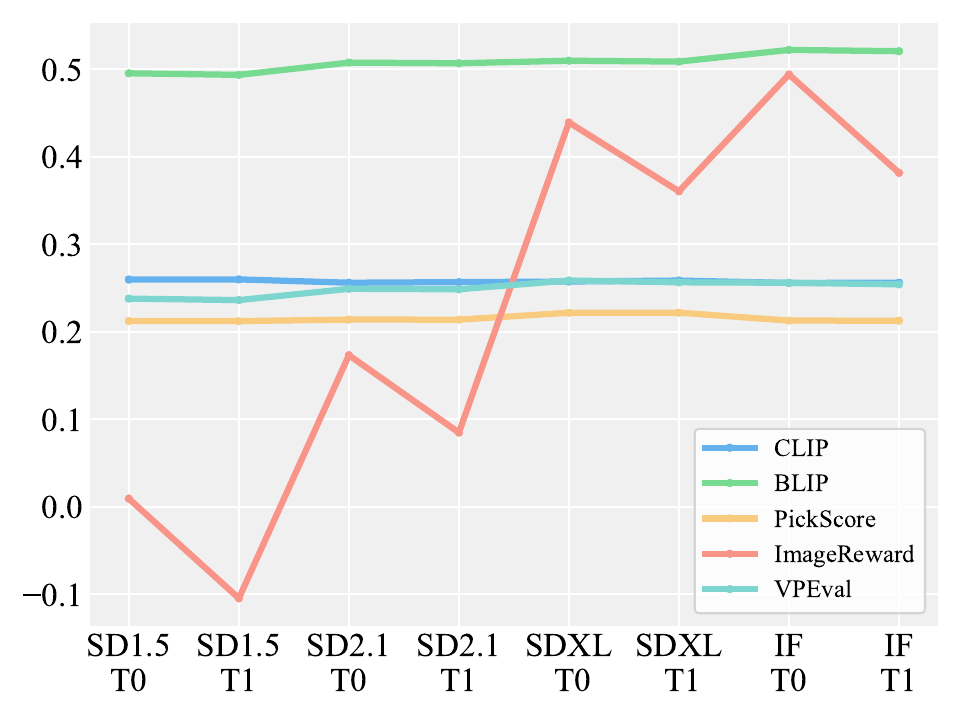}
        \label{fig:t2i_models_results_1}
   \end{subfigure}%
   \begin{subfigure}[b]{0.5\linewidth}
        \centering
        \includegraphics[width=\linewidth]{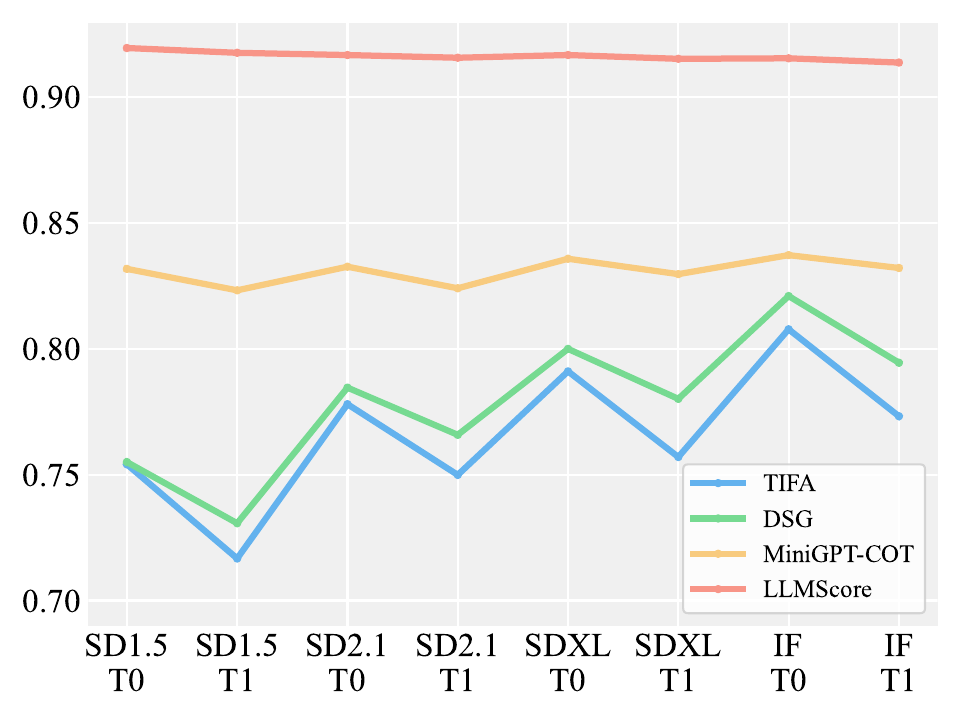}
        \label{fig:t2i_models_results_2}
   \end{subfigure}
   \vspace{-10mm}
   \caption{Evaluation results of Winoground-T2I $T^0$ and $T^1$ from 4 T2I models: SD1.5, SD2.1, SDXL, IF.}
   \label{fig:t2i_models_results}
\end{figure}

\begin{table}[htbp]
\centering

\begin{subtable}{\columnwidth}
\scriptsize
\centering
\setlength\tabcolsep{4pt} 
\begin{tabular}{ccc|cc|cc|cc}
\hline
\multirow{3}{*}{Metric} & \multicolumn{4}{c|}{SD1.5} & \multicolumn{4}{c}{SD2.1} \\ \cline{2-9} 
& \multicolumn{2}{c|}{$T_0$} & \multicolumn{2}{c|}{$T_1$} & \multicolumn{2}{c|}{$T_0$} & \multicolumn{2}{c}{$T_1$} \\ \cline{2-9} 
& $\tau (\uparrow)$ & $\rho (\uparrow)$ & $\tau (\uparrow) $ & $\rho (\uparrow)$ & $\tau (\uparrow)$ & $\rho (\uparrow)$ & $\tau (\uparrow) $ & $\rho (\uparrow)$ \\ \hline
CLIPScore    & 0.23 & 0.29 & 0.17 & 0.20 & 0.14 & 0.18 & 0.08 & 0.10 \\
BLIP-ITM     & 0.23 & 0.29 & 0.22 & 0.27 & 0.14 & 0.19 & 0.08 & 0.10 \\
PickScore    & 0.14 & 0.19 & 0.17 & 0.21 & 0.22 & 0.27 & 0.17 & 0.21 \\ 
ImageReward  & 0.30 & 0.37 & \underline{0.32} & \textbf{0.40} & 0.26 & \underline{0.34} & \underline{0.27} & \underline{0.34} \\ 
\hline
VPEval(fixed)   & 0.12 & 0.15 & 0.13 & 0.16 & 0.11 & 0.14 & 0.13 & 0.16 \\ 
VPEval(unfixed) & 0.12 & 0.15 & 0.10 & 0.12 & 0.12 & 0.16 & 0.14 & 0.17 \\ 
TIFA(fixed)     & 0.30 & 0.35 & 0.19 & 0.22 & 0.25 & 0.29 & 0.13 & 0.16 \\ 
TIFA(unfixed)   & 0.30 & 0.36 & 0.19 & 0.22 & \underline{0.27} & 0.31 & 0.12 & 0.14 \\
DSG(fixed)     & \textbf{0.35} & \textbf{0.41} & \underline{0.32} & \underline{0.38} & \underline{0.27} & 0.32 & 0.15 & 0.17 \\ 
DSG(unfixed)   & 0.34 & \underline{0.40} & 0.27 & 0.32 & \textbf{0.36} & \textbf{0.42} & 0.17 & 0.20 \\
\hline
MiniGPT4-CoT       & 0.05 & 0.06 & 0.11 & 0.12 & 0.15 & 0.17 & 0.04 & 0.05 \\
LLMScore      & 0.18 & 0.21 & -0.03 & -0.03 & -0.09 & -0.11 & 0.05 & 0.06 \\  
LLMScore$^+$     & 0.26 & 0.29 & 0.19 & 0.22 & 0.22 & 0.27 & 0.14 & 0.16 \\ 
LLMScore$^{++}$  & \underline{0.34} & \underline{0.40} & \textbf{0.33} & 0.37 & 0.22 & 0.27 & \textbf{0.31} & \textbf{0.36} \\ 
\hline
\end{tabular}
\caption{SD1.5 and SD2.1.
\label{tab:correlation_1_a}
}
\end{subtable}

\vspace{1em} 

\begin{subtable}{\columnwidth}
\scriptsize
\centering
\setlength\tabcolsep{4pt} 
\begin{tabular}{ccc|cc|cc|cc}
\hline
\multirow{3}{*}{Metric} & \multicolumn{4}{c|}{SDXL} & \multicolumn{4}{c}{IF} \\ \cline{2-9} 
& \multicolumn{2}{c|}{$T_0$} & \multicolumn{2}{c|}{$T_1$} & \multicolumn{2}{c|}{$T_0$} & \multicolumn{2}{c}{$T_1$} \\ \cline{2-9} 
& $\tau (\uparrow)$ & $\rho (\uparrow)$ & $\tau (\uparrow) $ & $\rho (\uparrow)$ & $\tau (\uparrow)$ & $\rho (\uparrow)$ & $\tau (\uparrow) $ & $\rho (\uparrow)$ \\ \hline
CLIPScore    & 0.13 & 0.17 & 0.14 & 0.18 & -0.03 & -0.04 & 0.11 & 0.13 \\
BLIP-ITM     & 0.14 & 0.18 & 0.17 & 0.22 & \underline{0.26} & \underline{0.33} & 0.15 & 0.20 \\
PickScore    & 0.20 & 0.25 & 0.25 & 0.31 & 0.19 & 0.24 & 0.19 & 0.24 \\ 
ImageReward  & 0.28 & 0.35 & 0.27 & \underline{0.34} & 0.26 & \underline{0.33} & \textbf{0.31} & \textbf{0.39} \\ 
\hline
VPEval(fixed)   & 0.01 & 0.02 & 0.02 & 0.03 & 0.08 & 0.10 & 0.12 & 0.16 \\ 
VPEval(unfixed) & 0.02 & 0.03 & 0.01 & 0.02 & 0.09 & 0.12 & 0.14 & 0.17 \\ 
TIFA(fixed)     & 0.21 & 0.25 & 0.14 & 0.17 & 0.25 & 0.30 & 0.19 & 0.23 \\ 
TIFA(unfixed)   & 0.20 & 0.24 & 0.15 & 0.18 & \underline{0.26} & 0.31 & 0.16 & 0.19 \\
DSG(fixed)     & \underline{0.33} & \underline{0.39} & \textbf{0.34} & \textbf{0.40} & 0.25 & 0.30 & \underline{0.25} & \underline{0.29} \\ 
DSG(unfixed)   & \textbf{0.34} & \textbf{0.41} & 0.28 & 0.33 & \textbf{0.30} & \textbf{0.36} & 0.21 & 0.25 \\
\hline
MiniGPT4-CoT       & 0.05 & 0.06 & 0.26 & 0.31 & 0.03 & 0.03 & 0.11 & 0.13 \\
LLMScore      & 0.17 & 0.20 & 0.19 & 0.21 & 0.04 & 0.04 & 0.03 & 0.03 \\  
LLMScore$^+$     & 0.26 & 0.30 & 0.12 & 0.14 & 0.23 & 0.26 &  0.05 & 0.06 \\ 
LLMScore$^{++}$  & 0.25 & 0.29 & \underline{0.29} & 0.33 & 0.24 & 0.28 & 0.12 & 0.14 \\ 
\hline
\end{tabular}
\caption{SDXL and IF.
\label{tab:correlation_1_b}
}
\end{subtable}

\caption{Results of inter-pair comparison on sample-100. The images are generated by SD1.5, SD2.1, SDXL and IF.}
\label{tab:correlation_1}

\end{table}

\begin{table}[htbp]
\centering

\begin{subtable}{\columnwidth}
\scriptsize
\centering
\setlength\tabcolsep{3pt} 
\begin{tabular}{ccc|ccc|ccccc}
\hline
\multirow{3}{*}{Metric} & \multicolumn{2}{c|}{$r_{same}$} & \multicolumn{3}{c|}{$p_{same}(\uparrow)$} & \multicolumn{5}{c}{$p_{diff}(\uparrow)$} \\ \cline{2-11}
 & \multirow{2}{*}{$\tau(\uparrow)$} & \multirow{2}{*}{$\rho(\uparrow)$} & $\epsilon=$ & $\epsilon=$ & $\epsilon=$ & $\Omega=$ & $\Omega=$ & $\Omega=$ & $\Omega=$ & $\Omega=$ \\ 
 & & & 0.05 & 0.07 & 0.1 & 0.4 & 0.5 & 0.6 & 0.7 & 0.8 \\ \hline
CLIPScore  & 0.44 & 0.59 & 0.26 & 0.35 & 0.44 & 0.06 & -- & -- & -- & -- \\ 
BLIP-ITM   & \underline{0.45} & \underline{0.61} & 0.17 & 0.24 & 0.31 & 0.09 & 0.13 & 0.20 & 0.38 & -- \\ 
PickScore  & 0.40 & 0.56 & 0.24 & 0.31 & 0.37 & 0.10 & -- & -- & -- & -- \\ 
ImageReward & 0.41 & 0.57 & 0.26 & 0.37 & 
0.44 & 0.07 & 0.10 & 0.10 & -- & -- \\  
\hline
VPEval(fixed) & 0.44 & 0.56 & \underline{0.41} & 0.44 & 0.52 & \textbf{0.80} & \textbf{0.80} & \textbf{0.89} & \textbf{1.00} & \textbf{1.00} \\
VPEval(unfixed) & \textbf{0.47} & \textbf{0.62} & \textbf{0.44} & \underline{0.50} & \underline{0.54} & 0.62 & 0.68 & 0.79 & \underline{0.83} & \textbf{1.00} \\ 
TIFA(fixed) & 0.41 & 0.53 & 0.31 & 0.39 & 0.39 & 0.51 & 0.52 & 0.60 & 0.69 & \underline{0.86} \\
TIFA(unfixed) & 0.40 & 0.54  & 0.30 & 0.35 & 0.37 & 0.47 & 0.51 & 0.58 & 0.68 & 0.77 \\
DSG(fixed) & \underline{0.45} & 0.57 & 0.37 & 0.44 & 0.46 & 0.52 & 0.53 & 0.59 & 0.72 & 0.73 \\
DSG(unfixed) & 0.40 & 0.56  & 0.24 & 0.28 & 0.35 & 0.36 & 0.38 & 0.47 & 0.55 & 0.71 \\
\hline
MiniGPT-CoT   & 0.12 & 0.15 & 0.28 & \textbf{0.56} & \textbf{0.56} & 0.38 & 0.39 & 0.48 & 0.57 & 0.71 \\ 
LLMScore & 0.15 & 0.20 & 0.19 & 0.20 & 0.20 & 0.43 & 0.46 & 0.53 & 0.80 & \textbf{1.00} \\ 
LLMScore$^+$ & 0.31 & 0.37 & 0.33 & 0.33 & 0.33 & 0.67 & 0.67 & \underline{0.82} & 0.75 & -- \\
LLMScore$^{++}$ & 0.10 & 0.15 & 0.24 & 0.24 & 0.24 & \underline{0.63} & \underline{0.73} & -- & -- & -- \\ 
\hline
\end{tabular}
\caption{SD1.5.
\label{tab:consistency_sd15}
}
\end{subtable}

\vspace{1em} 

\begin{subtable}{\columnwidth}
\scriptsize
\centering
\setlength\tabcolsep{3pt} 
\begin{tabular}{ccc|ccc|ccccc}
\hline
\multirow{3}{*}{Metric} & \multicolumn{2}{c|}{$r_{same}(\uparrow)$} & \multicolumn{3}{c|}{$p_{same}(\uparrow)$} & \multicolumn{5}{c}{$p_{diff}(\uparrow)$} \\ \cline{2-11}
 & \multirow{2}{*}{$\tau(\uparrow)$} & \multirow{2}{*}{$\rho(\uparrow)$} & $\epsilon=$ & $\epsilon=$ & $\epsilon=$ & $\Omega=$ & $\Omega=$ & $\Omega=$ & $\Omega=$ & $\Omega=$ \\ 
 & & & 0.05 & 0.07 & 0.1 & 0.4 & 0.5 & 0.6 & 0.7 & 0.8 \\ \hline
CLIPScore  & 0.44 & 0.62 & 0.23 & 0.37 & 0.40 & 0.08 & 0.06 & -- & -- & -- \\ 
BLIP-ITM   & 0.50 & 0.69 & 0.27 & 0.38 & 0.50 & 0.14 & 0.16 & 0.18 & 0.20 & -- \\ 
PickScore  & \textbf{0.55} & \textbf{0.73} & 0.19 & 0.29 & 0.54 & -- & -- & -- & -- & -- \\ 
ImageReward & 0.40 & 0.56 & 0.27 & 0.33 & 
0.42 & 0.13 & 0.15 & 0.19 & 0.24 & 0.35 \\  
\hline
VPEval(fixed) & 0.40 & 0.49 & \textbf{0.42} & \underline{0.48} & \underline{0.62} & \textbf{0.84} & \textbf{0.92} & \textbf{0.95} & \textbf{1.00} & \textbf{1.00} \\
VPEval(unfixed) & 0.43 & 0.54 & \textbf{0.42} & \textbf{0.58} & \textbf{0.65} & 0.26 & 0.38 & 0.53 & 0.57 & -- \\ 
TIFA(fixed) & 0.49 & 0.62 & 0.29 & 0.33 & 0.38 & 0.52 & 0.55 & 0.62 & 0.69 & 0.74 \\
TIFA(unfixed) & 0.51 & \underline{0.65} & 0.33 & 0.33 & 0.44 & 0.36 & 0.38 & 0.42 & 0.52 & 0.67 \\
DSG(fixed) & \underline{0.53} & 0.63 & \underline{0.40} & 0.44 & 0.54 & \underline{0.65} & \underline{0.67} & \underline{0.75} & \underline{0.78} & 0.80 \\
DSG(unfixed) & 0.43 & 0.54  & \underline{0.40} & 0.44 & 0.54 & 0.39 & 0.40 & 0.44 & 0.56 & 0.62 \\
\hline
MiniGPT-CoT   & -0.03 & -0.03 & 0.13 & 0.13 & 0.44 & 0.31 & 0.32 & 0.40 & 0.54 & \underline{0.83} \\ 
LLMScore & 0.11 & 0.13 & 0.23 & 0.25 & 0.25 & 0.39 & 0.40 & 0.41 & 0.48 & 0.60 \\ 
LLMScore$^+$ & 0.04 & 0.05 & 0.21 & 0.21 & 0.21 & -- & -- & -- & -- & -- \\
LLMScore$^{++}$ & 0.29 & 0.35 & 0.33 & 0.33 & 0.33 & 0.12 & 0.13 & 0.30 & -- & -- \\ 
\hline
\end{tabular}
\caption{SDXL.
\label{tab:consistency_sdxl}
}
\end{subtable}

\caption{Results of intra-pair comparison on sample-100. The images are generated by SD1.5 and SDXL.}
\label{tab:consistency}

\end{table}

\begin{table*}[ht]
\scriptsize
\centering
\setlength\tabcolsep{3pt} 
\begin{tabular}{ccc|cc|cc|cc|c|c|c}
\hline
\multirow{3}{*}{Metric} & \multicolumn{4}{c|}{SD1.5} & \multicolumn{4}{c|}{SDXL} & \multirow{3}{*}{Time(s)} & \multirow{3}{*}{$d(\cdot)$} & \multirow{3}{*}{$g(\cdot)$} \\ \cline{2-9}
 & \multicolumn{2}{c|}{$T_0$} & \multicolumn{2}{c|}{$T_1$} & \multicolumn{2}{c|}{$T_0$} & \multicolumn{2}{c|}{$T_1$} & & & \\ \cline{2-9}
 & $\bar \tau(\uparrow)$ & $\bar \rho(\uparrow)$ & $\bar \tau(\uparrow)$ & 
 $\bar \rho(\uparrow)$ & $\bar \tau(\uparrow)$ & $\bar \rho(\uparrow)$ & $\bar \tau(\uparrow)$ & $\bar \rho(\uparrow)$ & ($\downarrow$) & &  \\  
\hline
CLIPScore  & \textbf{1.00} & \textbf{1.00} & \textbf{1.00} & \textbf{1.00} & \textbf{1.00} & \textbf{1.00} & \textbf{1.00} & \textbf{1.00} & 0.04 &  embedding similarity & -- \\ 
BLIP-ITM   & \textbf{1.00} & \textbf{1.00} & \textbf{1.00} & \textbf{1.00} & \textbf{1.00} & \textbf{1.00} & \textbf{1.00} & \textbf{1.00} & 0.04 & embedding similarity & -- \\ 
PickScore  & \textbf{1.00} & \textbf{1.00} & \textbf{1.00} & \textbf{1.00} & \textbf{1.00} & \textbf{1.00} & \textbf{1.00} & \textbf{1.00} & 0.08 & embedding similarity & -- \\ 
ImageReward & \textbf{1.00} & \textbf{1.00} & \textbf{1.00} & \textbf{1.00} & \textbf{1.00} & \textbf{1.00} & \textbf{1.00} & \textbf{1.00} & 0.05 & embedding similarity & -- \\  
\hline
VPEval(fixed) & \textbf{1.00} & \textbf{1.00} & \textbf{1.00} & \textbf{1.00} & \underline{0.99} & \underline{0.99} & \underline{0.99} & \textbf{1.00} & 8.57 & pre-generated steps+ObjDet/VQA/OCR models & average \\
VPEval(unfixed) & 0.92 & 0.97 & 0.94 & 0.97 & 0.77 & 0.88 & 0.87 & 0.94 & 18.94 & LLM(step decomposition) + ObjDet/VQA/OCR & average \\
TIFA(fixed) & \textbf{1.00} & \textbf{1.00} & \textbf{1.00} & \textbf{1.00} & \textbf{1.00} & \textbf{1.00} & \textbf{1.00} & \textbf{1.00} & 1.17 & pre-generated question+VQA model & average \\
TIFA(unfixed) & \underline{0.97} & \underline{0.99} & \underline{0.97} & \underline{0.99} & 0.96 & 0.98 & 0.92 & \underline{0.95} & 5.85 & LLM(question generation) + VQA model & average \\ 
DSG(fixed) & \textbf{1.00} & \textbf{1.00} & \underline{0.97} & 0.98 & \textbf{1.00} & \textbf{1.00} & \textbf{1.00} & \textbf{1.00} & 1.08 & pre-generated question + VQA model & weighted average \\
DSG(unfixed) & 0.79 & 0.89 & 0.78 & 0.88 & 0.77 & 0.85 & 0.69 & 0.78 & 7.59 & LLM(parse) + LLM(question generation) + VQA model & weighted average \\
\hline
MiniGPT-CoT   & 0.05 & 0.06 & 0.05 & 0.05 & 0.10 & 0.11 & -0.03 & -0.04 & 34.10 & LLM(IC)+LLM(explanation) & LLM(rating) \\ 
LLMScore & 0.08 & 0.08 & 0.08 & 0.09 & 0.01 & 0.01 & 0.13 & 0.14 & 21.15 & ObjDet+IC+LLM(IC)+LLM(explanation) & LLM(rating) \\ 
LLMScore$^+$ & 0.26 & 0.26 & 0.34 & 0.35 & 0.29 & 0.29 & 0.31 & 0.32 & 15.69 & ObjDet+VLM(IC)+LLM(explanation) & LLM(rating) \\
LLMScore$^{++}$ & 0.28 & 0.31 & 0.18 & 0.19 & 0.18 & 0.20 & 0.19 & 0.21 & 24.83 & ObjDet+VLM(IC)+LLM(explanation)+LLM(verification) & LLM(rating) \\ 
\hline
\end{tabular}
\caption{Results of stability and efficiency on sample-100. 
The images are generated by SD1.5 and SDXL. In the $d(\cdot)$ of MiniGPT-CoT, the LLM refers to MiniGPT4 and in the $d(\cdot)$ of other metrics, the LLM refers to GPT-3.5. The VLM in $LLMScore^+$ and $LLMScore^{++}$ refers to instructblip. 
\label{tab:stability}
}
\end{table*}

\begin{figure*}[ht]
  \centering
   \includegraphics[width=1\linewidth]{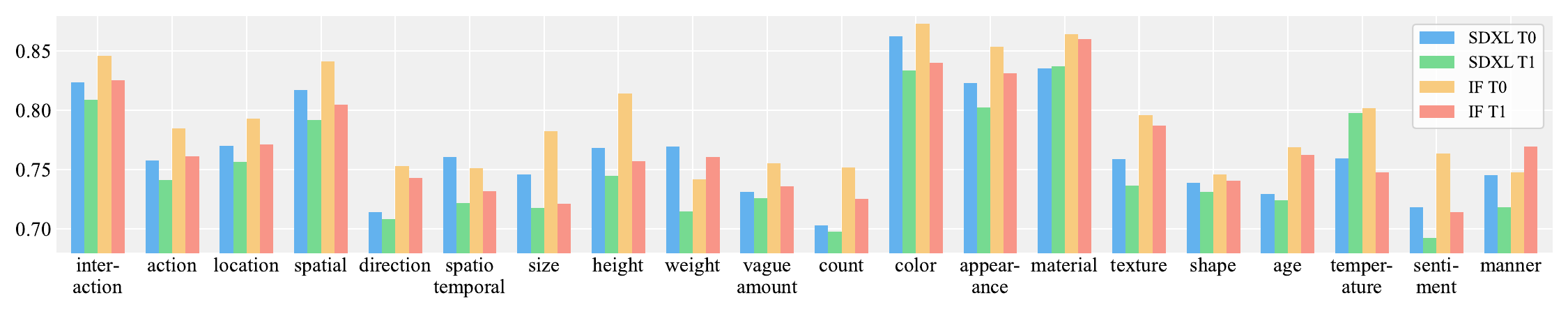}

   \caption{Evaluation results of categories of Winoground-T2I from SDXL and IF}
   \label{fig:categories_results}
\end{figure*}

\subsection{Experiment Results}
\noindent
\textbf{Evaluating Benchmarks}. We compare 7 benchmarks, each evaluated using 9 metrics on SDXL, as shown in Figure \ref{fig:benchmark_results}. Our benchmark scores are significantly lower than other benchmarks across all metrics, indicating the textual prompts are more challenging, except for BLIP-ITM, ImageReward and MiniGPT4-CoT.
%

Given three metrics, the VPEval-skill benchmark seems more challenging. Notably, 47\% of the prompts in VPEval-skill are single-object prompts, which should be theoretically depicted accurately and thus evaluated with high scores. However, in practice, this is not the case.
Upon analyzing the distribution of this benchmark and the results of the three metrics, we identify some key issues:  
1) ImageReward struggles to evaluate images in cartoon or aesthetic styles (e.g., SDXL), because it is trained for these models with the realistic style (e.g., SD 1.4). 
In an analysis of 100 cases from the \emph{Object} category, we observe that 62\% received negative ratings from ImageReward, yet the accuracy of these ratings is merely 6.5\%.
%
%
%
%
2) BLIP-ITM and MiniGPT4-CoT struggle to distinguish between text-based image evaluation and image-based text evaluation, resulting in lower scores. 
Because T2I models often over-interpret, adding unnecessary content to prompts that lack details, especially in single-object scenarios.

\noindent
\textbf{Evaluating T2I Models}. We compare the performance of T2I models on the simpler $T^0$ set and more challenging $T^1$ set in our benchmark, as shown in Figure \ref{fig:t2i_models_results}.
%
There are three trends observed from various metrics on SD1.5, SD2.1, SDXL and IF:
1) \emph{Continuous Performance Improvement}: The performance shows a consistently improvement from SD1.5 to IF in ImageReward, TIFA, and DSG metrics. 
The improvement suggests that the advanced text encoders of SDXL and IF enhance the understanding and generation of T2I models. As expected, models perform better on $T^0$ than $T^1$ in compositionality.
2) \emph{Slight Performance Improvement}: The performance shows a slight but perceptible improvement from SD1.5 to IF in BLIP, VPEval, and MiniGPT4-CoT metrics. The performance gap between $T^0$ and $T^1$ is also visible in MiniGPT4-CoT.
3) \emph{Minimal Performance Differences or Slight Decreases}: The performance of CLIP and PickScore shows negligible differences across models, while the performance of LLMScore shows a slight decline from SD1.5 to IF.

Our evaluation demonstrates significant divergence in performance using different metrics in the evaluations of benchmarks and T2I models, as shown in Figures \ref{fig:benchmark_results} and \ref{fig:t2i_models_results}. 
SDXL and IF, equipped with robust text encoders like CLIP-ViT-bigG, CLIP-ViT-L, and T5-XXL, exhibit superior compositionality in image generation than their predecessors. 
However, these advancements are not translated into significant improvements in most evaluation metrics, which seems a counterintuitive observation. 
%
The inconsistency evidences the need for a more reliable metric to evaluate the performance of T2I synthesis models accurately.

\noindent
\textbf{Evaluating Metrics}. We evaluate the reliability of various metrics on SD1.5 and SDXL from four different perspectives, as shown in Table \ref{tab:correlation_1}, Table \ref{tab:consistency} and Table \ref{tab:stability}. 

1) \emph{Inter-pair Comparison}. ImageReward, DSG, and LLMScore$^{++}$ stand out in their analogous metrics: feature based, visual programming based, and LLM’s CoT based. 
Specifically, based on BLIP, ImageReward aligns with human preferences through RLHF. 
%
DSG utilizes a VQA model to answer questions decomposed by LLMs and aggregate answer scores by semantics. 
This approach contrasts with VPEval's reliance on external tools and TIFA's method of averaging VQA scores.
LLMScore$^{++}$ improves image descriptions and reduces hallucinations through self-verification. 
These strategies collectively improve the reliability of the metrics.

2) \emph{Intra-pair Comparsion}. 
For similar human-rated pairs, feature based and visual programming based metrics show strong consistency, as shown in $r_{same}$ and $p_{same}$. 
For significantly different pairs, visual programming based metrics, especially VPEval, outperform others in nuanced analysis, as shown in $p_{diff}$. 
In contrast, feature based metrics show weak discriminative ability, while LLM’s CoT based metrics vary significantly, which warrants further analysis.

3) \emph{Stability}. 
%
Both feature based and visual programming based metrics are highly stable. The correlations of visual programming based metrics can approach 1 in repeated experiments when fixing these questions decomposed by LLMs.
However, LLM's CoT based metrics, which use LLMs in one step to examine the alignment of image and text, show serious irreproducibility.

4) \emph{Efficiency}. 
The evaluation can be broken down into distance functions $d(\cdot)$ and scoring functions $g(\cdot)$. 
Feature based metrics are most efficient, requiring no LLMs and no scoring function $g(\cdot)$. Visual programming based metrics use LLMs minimally for their distance function $d(\cdot)$, whereas LLM's CoT based metrics depend entirely on LLMs for both distance $d(\cdot)$ and scoring $g(\cdot)$ functions.

Therefore, we choose DSG to evaluate T2I performance across various categories. 
In inter-pair comparisons, DSG surpasses TIFA and VPEval, 
possibly due to the limited functionality of those external tools that VPEval used, while in intra-pair comparisons, VPEval surpasses DSG and TIFA, suggesting tool-based methods are superior within pairs.

\begin{figure*}[ht]
  \centering
   \includegraphics[width=1\linewidth]{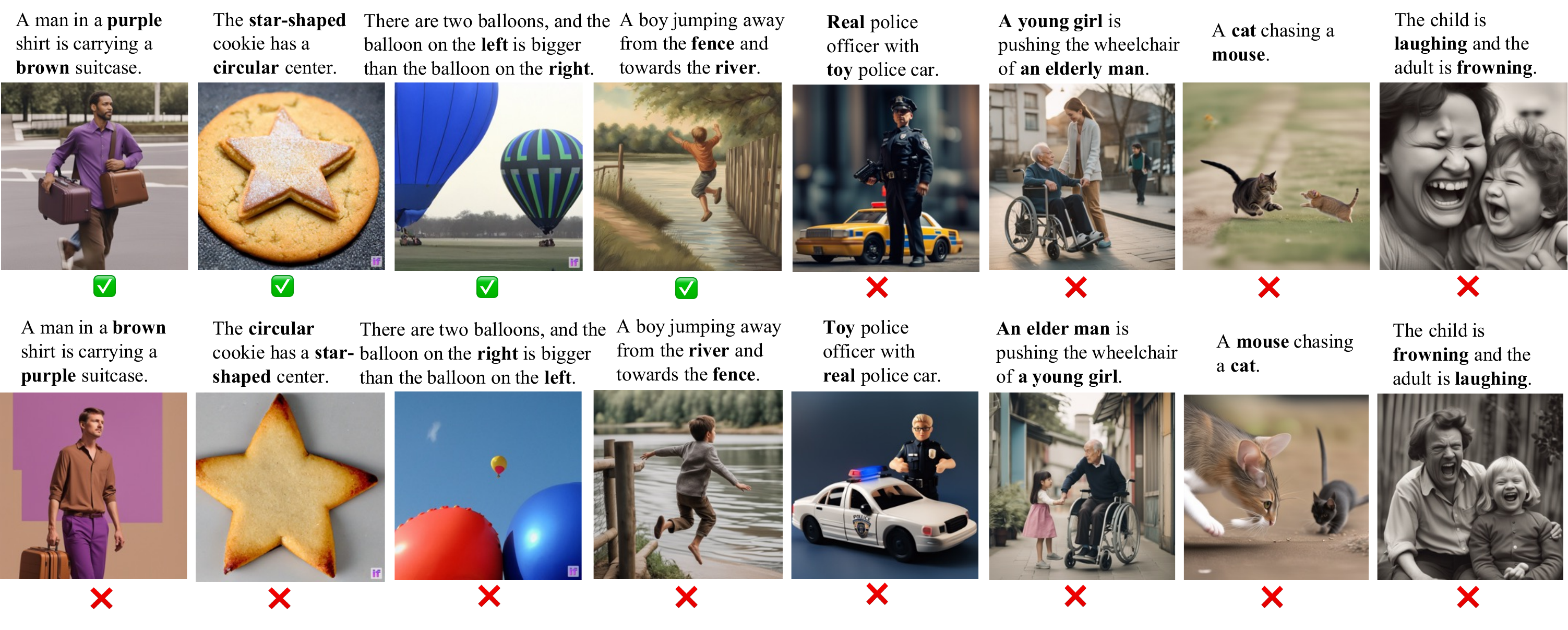}
   \caption{Success and Failure Cases in SDXL and IF}
   \label{fig:sd_xl_case}
\end{figure*}

\begin{figure*}[ht]
  \centering
   \includegraphics[width=1\linewidth]{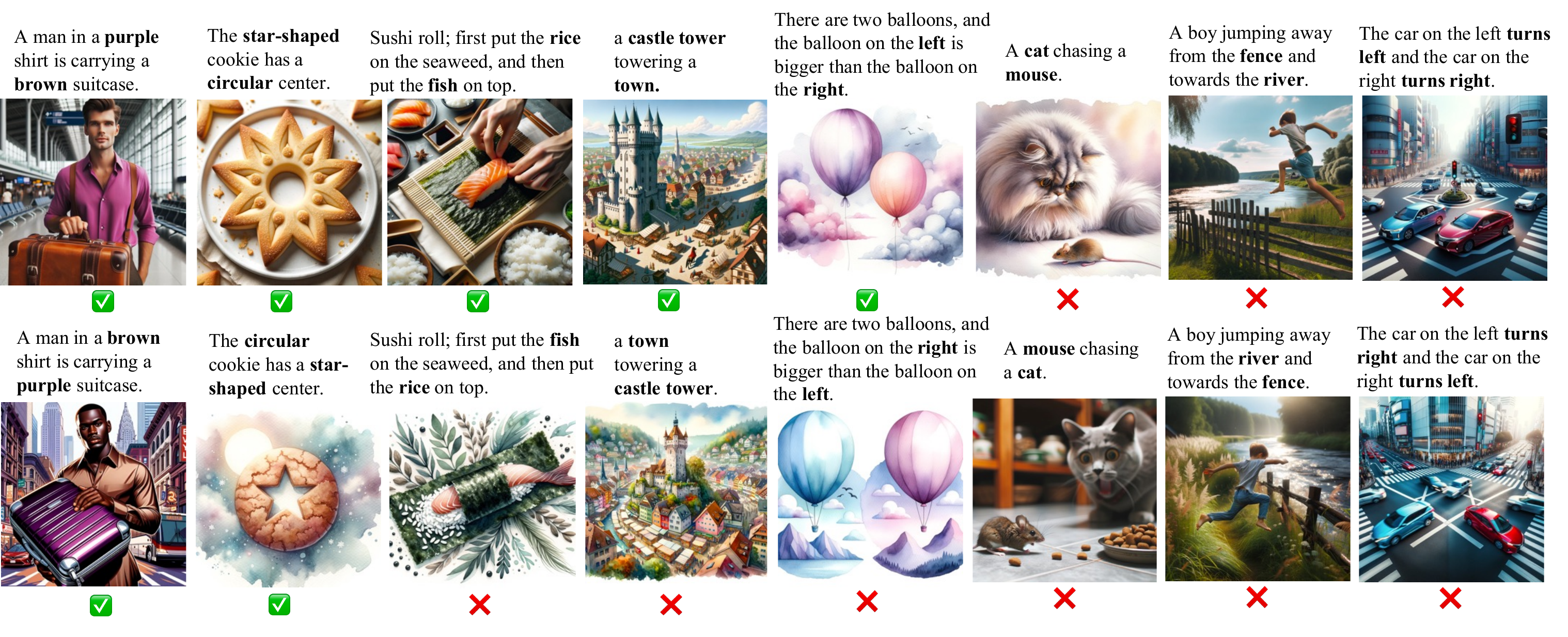}

   \caption{Success and Failure Cases in DALL-E 3}
   \label{fig:dalle_case}
\end{figure*}

\noindent
\textbf{Evaluating T2I models across Categories}. We evaluate the performance of SDXL and IF across categories, as shown in Figure \ref{fig:categories_results}. 
%
Those categories, which have been extensively discussed in other benchmarks, demonstrate strong generation capabilities, such as \emph{Color}, \emph{Material}, and \emph{Spatial}.
%
However, there is still room for improvement in accurately representing the following relations and attributes:
1) \emph{Relations}: \emph{Action, Location, Direction}, and \emph{Spatial-temporal}. 2) \emph{Attribute Comparisons}: \emph{Size, Height, Weight}, and \emph{Vague amount}. 3) \emph{Attributes Values}: \emph{Counting, Texture, Shape, Age, Temperature, Sentiment}, and \emph{Manner}. 

Based on the findings from our evaluation of SDXL and IF, it is apparent that challenges in T2I models extend beyond the issues of counting, layout, and attribute bounding. 
Our research identifies additional challenges in applying reasoning and knowledge within T2I synthesis, where transforming complex concepts into visual elements requires capabilities in several critical aspects.

1) \emph{Knowledge-Based Prompt Interpretation}, 
requires the understanding of specific factual knowledge or common sense for elements like action, direction, temperature, weight, and quantity,
such as the common sense that heavy objects sometimes correspond to a larger volume, and larger quantities are typically associated with a broader area.
For example, \emph{A cup of hot water in winter} may imply a cup with fog and steam on the sides and above the cup, while \emph{A boat sails away from the port towards the sea} may imply a boat with full sail heading seaward.
%

2) \emph{Reasoning-based Prompt Interpretation},
beyond factual knowledge, requires the reasoning of the sequence of events in a prompt to depict scenarios accurately in generated images.
For example, the sentence that \emph{the car was washed before someone drives it on a muddy road} implies a clean car on a muddy road, as it was washed beforehand. In contrast, the sentence that \emph{someone drove the car on a muddy road before the car is washed} implies a car covered in mud at a car wash, as it was driven on a muddy road before being washed.

3) \emph{Converting Non-Visual to Visual Elements}, 
requires converting time into spatial relationships or attributes, and translating dynamic actions into special visual techniques. For example, \emph{Sushi roll; first put the fish on the seaweed, then the rice on top} reveals the spatial relation of \emph{rice over seaweed}. Similarly, depicting motion in dynamic scenes, like \emph{a car stopping and a truck moving}, requires techniques used in photography and comics like multiple exposures or dynamic blurring to distinguish \emph{stopping} and \emph{moving}.

\subsection{Case Study}

\noindent
\textbf{Open Source T2I Models}. SDXL and IF succeed at the simpler $T^0$ level and fail at the $T^1$ level in the \emph{Attribute Comparisons} aspect, showing their difficulties with more complex attribute relationships. They also struggled with actions and direction, revealing weaknesses in these areas.

\noindent
\textbf{Commercial T2I Models}. DALL-E 3, utilizing the T5-XXL text encoder like IF, showed robust generative performance at both $T^0$ and $T^1$ for \emph{Attribute Values} and \emph{Attribute Comparison}. 
%
Although DALL-E 3 emphasizes the significance of better captions, advanced text encoders also play a crucial role in understanding complex compositional categories.
However, DALL-E 3 faces challenges at the $T^1$ level in more complex categories such as \emph{Actions, Diction, Spatial-temporal}, and fine-grained \emph{Attribute Composition}.

\section{Conclusion}
In this paper, we introduce Winoground-T2I, a benchmark for evaluating the compositionality of T2I synthesis models, featuring contrastive sentence pairs across 20 categories. We propose a strategy for selecting a reliable metric from four perspectives to ensure effective and persuasive evaluation. Based on this more reliable metric and our contrastive sentences benchmark, we analyze the strengths and weaknesses through statistical and case studies of current T2I models.
{
    \small
    \bibliographystyle{ieeenat_fullname}
    \bibliography{main}
}

\clearpage
\setcounter{page}{1}
\maketitlesupplementary

\section{Additional Related Work}

\noindent
\textbf{Winoground Dataset}. Winoground~\cite{DBLP:conf/cvpr/ThrushJBSWKR22} is a manually annotated dataset designed to evaluate the compositional reasoning of visual-language models (VLMs), particularly their sensitivity to word order~\cite{DBLP:conf/iclr/Yuksekgonul0KJ023,DBLP:conf/cvpr/MaHGGGK23,DBLP:journals/corr/abs-2303-17590}. The dataset contains 400 sentence pairs, half identified as focusing on compositionality~\cite{DBLP:conf/emnlp/DiwanBCHM22}. It shows promise for evaluating reasoning capabilities in T2I models~\cite{DBLP:journals/corr/abs-2204-13807,DBLP:journals/corr/abs-2311-04287}. However, applying it in T2I synthesis evaluation faces challenges due to ambiguous sentences, which, despite being suitable for VLMs, often lack the detailed information necessary for T2I synthesis. Additionally, its limited scale and linguistic-based categorization also present constraints.

%

\section{Details of Winoground-T2I Construction}

\subsection{Template Acquisition}

We manually extract templates and modification rules from 200 compositional cases in Winoground~\cite{DBLP:conf/cvpr/ThrushJBSWKR22}, labeled as \emph{no-tag} in subsequent research\cite{DBLP:conf/emnlp/DiwanBCHM22}. These examples are illustrated in Table ~\ref{tab:template}. We open source these templates alongside our benchmark.

\begin{table*}[ht]
\centering
\footnotesize
\setlength\tabcolsep{6pt} 
\begin{tabular}{lll}
\specialrule{.1em}{.05em}{.05em} 
\textbf{Seed Sentence Pairs from Winoground} & \textbf{Templates} & \textbf{Modification Rule} \\
\hline

\begin{tabular}[c]{@{}l@{}}$caption\_0$: a bird eats a snake\\$caption\_1$: a snake eats a bird\end{tabular} & \begin{tabular}[c]{@{}l@{}}$T^0$: [Object1] [Interact] [Object2]\\$T^1$: [Object2] [Interact] [Object1]\end{tabular} & [Object1] $\leftrightarrow$ [Object2] \\ \hline

\begin{tabular}[c]{@{}l@{}}$caption\_0$: a person is in a helicopter which is in a car\\$caption\_1$: a person is in a car which is in a helicopter\end{tabular} & \begin{tabular}[c]{@{}l@{}}$T^0$: [Object1] [Action1] [Location1] which is in [Location2]\\$T^1$: [Object1] [Action1] [Location2] which is in [Location1]\end{tabular} & [Location1] $\leftrightarrow$ [Location2] \\ \hline

\begin{tabular}[c]{@{}l@{}}$caption\_0$: there are some pineapples in boxes, and\\far more pineapples than boxes\\$caption\_1$: there are some boxes containing pineapples,\\and far more boxes than pineapples\end{tabular} & \begin{tabular}[c]{@{}l@{}}$T^0$: ([Location], )(There be)[Object1] [locate in] [Object2], \\and far more [Object1] than [Object2]\\$T^1$: ([Location], )(There be)[Object2] [contain] [Object1], \\and far more [Object2] than [Object1]\end{tabular} & [Object1] $\leftrightarrow$ [Object2] \\ \hline

\begin{tabular}[c]{@{}l@{}}$caption\_0$: the person sitting down is supporting the\\ person standing up\\$caption\_1$: the person standing up is supporting the\\person sitting down\end{tabular} & \begin{tabular}[c]{@{}l@{}}$T^0$: [Object1] [Action1] [Interact] [Object2] [Action2]\\$T^1$: [Object1] [Action2] [Interact] [Object2] [Action1]\end{tabular} & [Action1] $\leftrightarrow$ [Action2] \\ \hline

\begin{tabular}[c]{@{}l@{}}$caption\_0$: the person with green legs is running quite\\ slowly and the red legged one runs faster\\$caption\_1$: the person with green legs is running faster\\and the red legged one runs quite slowly\end{tabular} & \begin{tabular}[c]{@{}l@{}}$T^0$: [Object1] [Appearance1] [Action1] slowly and [Object2] \\ \relax [Appearance2] [Action2] faster\\$T^1$: [Object1] [Appearance1] [Action1] faster and [Object2] \\ \relax [Appearance2] [Action2] slowly\end{tabular} & slowly $\leftrightarrow$ faster \\


\specialrule{.1em}{.05em}{.05em} 
\end{tabular}
\caption{\label{tab:template}
Examples of extracted templates and modification rules between templates.
}
\end{table*}

\subsection{Contrastive Sentences Generation}
We have details of the prompts that GPT3.5 uses for generating contrastive sentences. The seed pairs are $caption\_0$ and $caption\_1$, as illustrated in Table \ref{tab:template}.

\noindent
\textbf{Template-guided Generation}. The prompt for generating the $T^0$ guided by the templates and seed pairs is: \emph{Assuming you are a linguist, you can create a similar sentence following the structure of given sentences. The two sentences are [slot for $caption\_0$] and [slot for $caption\_1$]. Their structures are both [slot for the template of $T^0$ and $T^1$]. Please create a similar [slot for the template of $T^0$ and $T^1$] sentence, and diversify your sentence as much as possible by using different themes, scenes, objects, predicates, verbs, and modifiers. } 

\noindent
\textbf{Contrastive Pairs Acquisition}. The prompt for generating the $T^1$ based on the $T^0$ and its modification rule is: \emph{Assuming you are a linguist, you can judge the structure of existing sentences and imitate more new sentences with similar structure but varied content. Step 1: Input some [slot for the template of $T^0$ and $T^1$] sentences. We call each sentence as "text0". Step 2: For each "text0", perform the change, which is [slot for the modification rule], and keep the other words unchanged as "text1". For example, text0=[slot for $caption\_0$]. Only swap/move [slot for the modification rule] and keep the other words unchanged to generate text1=[slot for $caption\_1$].}

\subsection{Quality Control Criteria}

\begin{table*}[ht]
\footnotesize
\centering
\setlength\tabcolsep{6pt} 
\begin{tabular}{llll}
\specialrule{.1em}{.05em}{.05em} 
\textbf{Type}                 & \textbf{\begin{tabular}[c]{@{}l@{}}Valid Criteria\end{tabular}} & \textbf{$T^0$} & \textbf{$T^1$} \\ \hline
\multirow{3}{*}{Basic} & \begin{tabular}[c]{@{}l@{}}Complete Expression\end{tabular} & \begin{tabular}[c]{@{}l@{}}Swinging on the swing and off the metal chains.\end{tabular} & \begin{tabular}[c]{@{}l@{}}Swinging off the swing and on the metal chains.\end{tabular} \\ \cline{2-4} 
& \begin{tabular}[c]{@{}l@{}}Clear and Concrete Objects\end{tabular} & \begin{tabular}[c]{@{}l@{}}A brighter sun is shining on a dimmer object.\end{tabular} & \begin{tabular}[c]{@{}l@{}}A dimmer sun is shining on a brighter object.\end{tabular} \\ \cline{2-4} 
& \begin{tabular}[c]{@{}l@{}}Reasonable Semantics\end{tabular} & \begin{tabular}[c]{@{}l@{}}An engineer builds a bridge.\end{tabular} & \begin{tabular}[c]{@{}l@{}}A bridge builds an engineer.\end{tabular} \\ \hline

\multirow{4}{*}{Visualizable} & \begin{tabular}[c]{@{}l@{}}Visually Depicted Elements\end{tabular} & \begin{tabular}[c]{@{}l@{}}There are more salads than burgers on the menu.\end{tabular} & \begin{tabular}[c]{@{}l@{}}There are more burgers than salads on the menu.\end{tabular} \\ \cline{2-4} 
& \begin{tabular}[c]{@{}l@{}}Static Scene or\\Multiple Exposure Scene\end{tabular} & \begin{tabular}[c]{@{}l@{}}The wave is moving faster and the fish is \\swimming slowly.\end{tabular} & \begin{tabular}[c]{@{}l@{}}The fish is swimming faster and the wave is \\moving slowly.\end{tabular} \\ \cline{2-4} 
& \begin{tabular}[c]{@{}l@{}}Moderate Details\end{tabular} & \begin{tabular}[c]{@{}l@{}}In the library, there are a stack of books and\\some more magazines. \\ \end{tabular} & \begin{tabular}[c]{@{}l@{}}In the library, there are a stack of magazine and\\some more books.\\ \end{tabular} \\ \cline{2-4} 
& \begin{tabular}[c]{@{}l@{}}Quantifiable Comparison\end{tabular} & \begin{tabular}[c]{@{}l@{}}There are more ants than bees in the garden.\end{tabular} & \begin{tabular}[c]{@{}l@{}}There are more bees than ants in the garden.\end{tabular} \\ \hline

\multirow{3}{*}{Contrastive}  & \begin{tabular}[c]{@{}l@{}}Modification Rules\end{tabular} & \begin{tabular}[c]{@{}l@{}}A sharp knife is on a dull cutting board.\end{tabular} & \begin{tabular}[c]{@{}l@{}}A dull cutting board is under a sharp knife.\end{tabular} \\ \cline{2-4} 
& \begin{tabular}[c]{@{}l@{}}Distinct Textual Semantics\end{tabular} & \begin{tabular}[c]{@{}l@{}}The boat is on the dock and the fisherman is on\\the pier.\end{tabular} & \begin{tabular}[c]{@{}l@{}}The boat is on the pier and the fisherman is on\\the dock.\end{tabular} \\ \cline{2-4} 
& \begin{tabular}[c]{@{}l@{}}Visually Distinguishable\end{tabular} & \begin{tabular}[c]{@{}l@{}}There's a delicious chocolate cake with a bitter\\ coffee frosting.\end{tabular} & \begin{tabular}[c]{@{}l@{}}There's a bitter chocolate cake with a delicious\\coffee frosting.\end{tabular} \\ \hline

\multirow{4}{*}{Recognizable} & \begin{tabular}[c]{@{}l@{}}Item-Specific Scene\end{tabular} & \begin{tabular}[c]{@{}l@{}}There are more books than shelves in this library.\end{tabular} & \begin{tabular}[c]{@{}l@{}}There are more shelves than books in this library.\end{tabular} \\ \cline{2-4} 
& \begin{tabular}[c]{@{}l@{}}Item-Specific Character\end{tabular} & \begin{tabular}[c]{@{}l@{}}A photographer wearing a camera strap with his\\lens in the air and a videographer wearing a tripod.\end{tabular} & \begin{tabular}[c]{@{}l@{}}A photographer wearing a tripod with his lens in\\the air and a videographer wearing a camera strap.\end{tabular}  \\ \cline{2-4} 
& \begin{tabular}[c]{@{}l@{}}Attire-based Character\end{tabular} & \begin{tabular}[c]{@{}l@{}}The soldier in the barracks is cleaning equipment\\and the officer in the office is reviewing reports.\end{tabular} & \begin{tabular}[c]{@{}l@{}}The soldier in the barracks is reviewing reports and \\the officer in the office is cleaning equipment.\end{tabular}\\ \cline{2-4} 
& \begin{tabular}[c]{@{}l@{}}Action-based Character\end{tabular} & \begin{tabular}[c]{@{}l@{}}The businessman is wearing navy suit and red tie.\end{tabular} & \begin{tabular}[c]{@{}l@{}}The businessman is wearing red suit and navy tie.\end{tabular} \\ 

\specialrule{.1em}{.05em}{.05em} 
\end{tabular}
\caption{
Examples of LLM-generated sentence pairs ($T^0$, $T^1$) which violate the criteria. 
\label{tab:invalid_samples}
}
\end{table*}

\begin{table*}[ht]
\centering
\footnotesize
\setlength\tabcolsep{6pt} 
\begin{tabular}{llll}
\specialrule{.1em}{.05em}{.05em} 
\textbf{Aspect} & \textbf{Category} & \textbf{$T^0$} & \textbf{$T^1$} \\
\hline
\multirow{6}{*}{Relation} & Action & A dog sits and a cat stands. & A dog stands and a cat sits. \\ \cline{2-4} 
                  & Interaction & \begin{tabular}[c]{@{}l@{}}An old person kisses a young person.\end{tabular} & \begin{tabular}[c]{@{}l@{}}A young person kisses an old person.\end{tabular} \\ \cline{2-4} 
                  & Location & \begin{tabular}[c]{@{}l@{}}The soft teddy bear is on the bed and the hard toy car\\is on the shelf.\end{tabular} & \begin{tabular}[c]{@{}l@{}}The soft teddy bear is on the shelf and the hard toy car\\is on the bed.\end{tabular}\\ \cline{2-4} 
                  & Spatial & \begin{tabular}[c]{@{}l@{}}A green apple sits atop a red leaf. \end{tabular} & \begin{tabular}[c]{@{}l@{}}A red leaf sits atop a green apple. \end{tabular} \\ \cline{2-4} 
                  & \begin{tabular}[c]{@{}l@{}}Spatial-Temporal\end{tabular} & \begin{tabular}[c]{@{}l@{}}Sushi roll; first put the fish on the seaweed, and then\\put the rice on top.\end{tabular} & \begin{tabular}[c]{@{}l@{}}Sushi roll; first put the rice on the seaweed, and then \\put the fish on top.\end{tabular} \\ \cline{2-4}
                  & Direction & \begin{tabular}[c]{@{}l@{}}A boy jumps away from the fence and towards the river.\end{tabular} & \begin{tabular}[c]{@{}l@{}}A boy jumps away from the river and towards the fence.\end{tabular} \\ \cline{2-4} 
\hline
\multirow{4}{*}{\begin{tabular}[c]{@{}l@{}}Attribute\\Comparison\end{tabular}} & Scale & \begin{tabular}[c]{@{}l@{}}The cake and the plate; the cake is too big for the plate.\end{tabular} & \begin{tabular}[c]{@{}l@{}}The cake and the plate; the plate is too big for the cake.\end{tabular} \\ \cline{2-4}
                  & Height & \begin{tabular}[c]{@{}l@{}}A dinosaur towering over a human.\end{tabular} & \begin{tabular}[c]{@{}l@{}}A human towering over a dinosaur.\end{tabular} \\ \cline{2-4}
                  & Weight & \begin{tabular}[c]{@{}l@{}}The athlete with a heavy backpack is walking quite\\slowly and the one with a light bag is running faster.\end{tabular} & \begin{tabular}[c]{@{}l@{}}The athlete with a light backpack is walking quite\\slowly and the one with a heavy bag is running faster.\end{tabular} \\ \cline{2-4}
                  & \begin{tabular}[c]{@{}l@{}}Vague Amount\end{tabular} & \begin{tabular}[c]{@{}l@{}}A cake with more frosting on the top than on the slides.\end{tabular} & \begin{tabular}[c]{@{}l@{}}A cake with more frosting on the slides than on the top.\end{tabular} \\ \cline{2-4}
\hline
\multirow{10}{*}{\begin{tabular}[c]{@{}l@{}}Attribute\\Values\end{tabular}} & Color & \begin{tabular}[c]{@{}l@{}}A man in a purple shirt is carrying a brown suitcase.\end{tabular} & \begin{tabular}[c]{@{}l@{}}A man in a brown shirt is carrying a purple suitcase.\end{tabular} \\ \cline{2-4}
                  & Counting & \begin{tabular}[c]{@{}l@{}}Four dogs in a doghouse and one dog barking outside.\end{tabular} & \begin{tabular}[c]{@{}l@{}}One dogs in a doghouse and four dog barking outside.\end{tabular} \\ \cline{2-4}
                  & Texture & \begin{tabular}[c]{@{}l@{}}Two fish; the one in the tank has stripes and the one in \\the bowl  doesn't.\end{tabular} & \begin{tabular}[c]{@{}l@{}}Two fish; the one in the bowl has stripes and the one in \\the tank  doesn't.\end{tabular}\\ \cline{2-4}
                  & Material & \begin{tabular}[c]{@{}l@{}}There's a satin teddy bear with a furry bow.\end{tabular} & \begin{tabular}[c]{@{}l@{}}There's a furry teddy bear with a satin bow.\end{tabular} \\ \cline{2-4}
                  & Shape & \begin{tabular}[c]{@{}l@{}}The circular suitcase has an oblong lock.\end{tabular} & \begin{tabular}[c]{@{}l@{}}The oblong suitcase has an circular lock.\end{tabular} \\ \cline{2-4}
                  & Age & \begin{tabular}[c]{@{}l@{}}The person on the left is old and the person on the\\right is young.\end{tabular} & \begin{tabular}[c]{@{}l@{}}The person on the right is old and the person on the\\left is young.\end{tabular} \\ \cline{2-4}
                  & Sentiment & \begin{tabular}[c]{@{}l@{}}The happy child is playing next to a sad clown.\end{tabular} & \begin{tabular}[c]{@{}l@{}}The sad child is playing next to a happy clown.\end{tabular} \\ \cline{2-4}
                  & Temperature & \begin{tabular}[c]{@{}l@{}}Iced coffee and steaming tea.\end{tabular} & \begin{tabular}[c]{@{}l@{}}Steaming coffee and iced tea.\end{tabular}\\ \cline{2-4}
                  & Manner & \begin{tabular}[c]{@{}l@{}}The building on the corner has a modern design and\\the monument in the park has a classic design.\end{tabular} & \begin{tabular}[c]{@{}l@{}}The building on the corner has a classic design and\\the monument in the park has a modern design.\end{tabular} \\ \cline{2-4}
                  & Appearance & \begin{tabular}[c]{@{}l@{}} The boy with a blue shirt has long hair and the girl\\in the pink dress has short hair.\end{tabular} & \begin{tabular}[c]{@{}l@{}} The boy with a blue shirt has short hair and the girl\\in the pink dress has long hair.\end{tabular}  \\ 
\specialrule{.1em}{.05em}{.05em} 
\end{tabular}
\caption{\label{tab:table_category}
Examples of categories in Winoground-T2I.
}
\end{table*}

The \emph{valid} input text for T2I models are defined by 14 criteria, examples in Table ~\ref{tab:invalid_samples}.
\begin{itemize}
    \item \emph{Basic}
    \begin{itemize}
        \item \emph{Complete Expression:} Both sentences should be complete and free from obvious linguistic errors.

        \item \emph{Clear and Concrete Objects }: Both sentences must be clear and unambiguous, contextually or inherently, and specifically describe tangible objects, steering clear of abstract concepts.

        \item \emph{Reasonable Semantics:} Both sentences must maintain logical coherence in their respective contexts. 
        The reasonable definition includes real-world plausibility or scenarios typically seen as implausible in virtual or imaginative settings (like children's literature, animations, or science fiction), such as flying pigs or dinosaurs piloting planes. 
        For example, \emph{a shorter person can reach a higher shelf while a taller one cannot} is not reasonable in any world.

        
        
    \end{itemize}

    \item \emph{Visualizable}
    \begin{itemize}
        \item \emph{Visually Depicted Element:} 
        Both sentences must 
        convey visual elements, including objects, scenes, actions, and attributes, ensuring that the text prompts are visually depictable and the image content is identifiable during evaluation.

        \item \emph{Static Scene or Multiple Exposure Scene: }
        Both sentences should be visually representable through images alone, negating the need for video, audio, or other sensory inputs like touch and smell. Temporal aspects, procedures, and comparisons in test cases must be conveyable within a single image's scope.

        \item \emph{A Moderate Level of Details: } Sentences should maintain a moderate level of detail with similar scales for objects and scenes. Excessive or mismatched scales can result in sentences that are challenging to depict. 
        For example, comparing the quantity of books and magazines \emph{in a library} is less suitable than \emph{on a table}.

        \item \emph{Quantifiable Comparison: } Comparisons in both sentences should be quantifiable, using measures like counts, areas, or volumes.
        For example, \emph{There are more students in the classroom than words on the blackboard} are difficult to compare quantitatively.
        
    \end{itemize}

    \item \emph{Contrastive}
    \begin{itemize}
        \item \emph{Following Modification Rules:} Generated samples must strictly follow the designated manual template, including word swapping and moving.
        \item \emph{Distinct Textual Semantics}: Two sentences must have distinct textual semantics. Otherwise, the pairs are considered invalid.

        \item \emph{Visually Distinguishable:} 
        Two sentences should be visually distinct, with clear differentiation regarding the visual characteristics of the objects or scenes described. 
        Subtle differences requiring very close observation are not considered \emph{distinct visual differences}.
        
    \end{itemize}

    \item \emph{Recognizable}
    \begin{itemize}


        \item \emph{Item-Specific Scenes}: Scenes in sentences should be identifiable, maintaining key elements for recognition. Otherwise, identification may be challenging. For instance, a sentence describing a \emph{library} where \emph{bookshelves outnumber books} might be unrecognizable, as we typically expect a library to contain many books.
         

        \item \emph{Item-Specific Characters}: When a sentence depicts a character through associations with specific items, these items or behaviors should remain consistent for easy identification. If not, the character may be hard to recognize. For instance, \emph{chefs} are usually associated with \emph{chef's attire, cooking utensils, and kitchens}.
        
        \item \emph{Attire-Based Characters:} When a sentence presents characters identifiable by their attire, such as \emph{firefighters, police officers, soldiers, doctors, and nurses}, their clothing should remain consistent for clear recognition. Changes in attire could obscure their identities.
        
        \item \emph{Action-Based Characters:} When a sentence features characters defined by specific actions or interactions, such as bartenders (mixing drinks), businessmen (negotiating), journalists (interviewing), divers (deep-sea diving), their typical activities should be consistent. Altering distinctive features or placing characters in unusual scenarios may obscure their identities.

    \end{itemize}
    
\end{itemize}

\subsection{Categories}

The sentences in Winoground-T2I are categorized into 3 aspects and 20 categories based on their compositional semantics, details in Table \ref{tab:table_category}.
These categories are \emph{Relation}, \emph{Attribute Comparison} and \emph{Attribute Values}. Specifically, \emph{Relation} includes 6 categories: \emph{Action, Interaction, Location, Spatial, Spatial-Temporal, Direction}. \emph{Attribute Contrast} includes 4 categories: \emph{Scale, Height, Weight, Vague Amount}. \emph{Attribute Values} includes 10 categories: \emph{Color, Counting, Texture, Material, Shape, Age, Sentiment, Temperature, Manner}, and \emph{Appearance}.

\section{Details of Experiment Setting}

\subsection{T2I Synthesis Models}

We generate one image using the following mainstream T2I diffusion models. The schedulers in SD1.5 and SD2.1 are both set to DPM-Solver++. All other settings are as default.
\begin{itemize}
    \item \textbf{Stable Diffusion v1.5}: A widely used version of Stable Diffusion, denoted as SD 1.5. It uses CLIP ViT-L as the text encoder, U-Net to generate latent images, and VAE to convert images into latent features.
    \item \textbf{Stable Diffusion v2.1}: SD 2.1, a significant improvement over SD 1.5 in image quality, supports larger resolution images, such as $512 \times 512$ and $768 \times 768$. SD 2.1 retains the same text encoder as SD 1.5.
    \item \textbf{Stable Diffusion XL v1.0}: SDXL 1.0, comparable in quality to Midjourney, enhances the three main components of SD: U-Net, VAE, and CLIP Text Encoder. Compared to the previous versions, Stable Diffusion XL employs two CLIP Text Encoders: CLIP ViT-bigG (text encoder 694.7M parameters) and CLIP ViT-L (text encoder 123.65M parameters), significantly enhancing SDXL's ability to understand and comprehend text. 
    \item \textbf{DeepFloyd IF}: DeepFloyd IF, denoted as DF-IF, is an open-source version of Imagen, which utilizes a T5-XXL to encode text and three different diffusion models to achieve image generation with high resolution. Compared to the text encoder in CLIP, T5-XXL features 4.6 billion parameters, enabling it to carry richer semantic information. This much larger size helps overcome CLIP's insensitivity to text order.

\end{itemize}

\subsection{Evaluation Metrics}

We evaluate several metrics for reliability, including feature-based, visual programming, and LLM's CoT-based metrics. Additionally, inspired by studies indicating that self-validation in large models enhances result reliability~\cite{DBLP:journals/corr/abs-2309-11495,DBLP:journals/corr/abs-2306-03872}, we refine LLMScore to determine if improved image descriptions and self-verification augment reliability, as detailed in LLMScore$^+$ and LLMScore$^{++}$.

\begin{itemize}
    \item Feature Based Metrics:
    \begin{itemize}
        \item \textbf{CLIPScore} is the cosine similarity between image and text features extracted from ViT-L/14.
        \item \textbf{BLIP-ITM} is the cosine similarity between image and text features extracted from BLIP2. 
        \item \textbf{PickScore} is a CLIP-style model trained with a variant of InstructGPT’s reward model objective. 
        \item \textbf{ImageReward} is a BLIP-style model
        trained by a variant of InstructGPT’s reward model objectives. 
    \end{itemize}
    \item Visual Programming Based Metrics: 
    \begin{itemize}
        \item \textbf{VPEval} evaluates the accuracy of an image generated from text using external tools, such as object detector and OCR models, which are called upon by steps generated by prompting LLMs (GPT-3.5). Considering the inherent uncertainty in GPT-3.5 responses, we compare pre-generated \emph{fixed steps} for uniform evaluation with \emph{unfixed steps} for zero-shot evaluation.
        \item \textbf{TIFA} evaluates the accuracy of an image generated from text using VQA by answering questions based on the generated image. Questions are produced by prompting LLMs (GPT-3.5) and filtered by a model. Considering the inherent uncertainty in GPT-3.5 responses, we compare pre-generated \emph{fixed questions} for uniform evaluation with \emph{unfixed questions} for zero-shot evaluation.
        
        \item \textbf{DSG} is an enhanced version of TIFA, emphasizing a sentence's key semantic components. These semantics, including \emph{Entity}, \emph{Relation}, and \emph{Attribute}, are also extracted by LLMs. The weight of question responses is adjusted based on the significance of these components. Other aspects of DSG align with the original TIFA settings.
        
    \end{itemize}

    \item LLM's CoT Based Metrics: 
    \begin{itemize}
        \item \textbf{MiniGPT-CoT} utilize MiniGPT-4 (based on vicuna-13b) to evaluate the T2I fidelity. It first generates a long and comprehensive image caption and then evaluates the alignment between this description and the text prompt.
        \item \textbf{LLMScore} utilize a LLM (we use GPT-3.5) to evaluate the T2I fidelity. It first extends a short caption with bounding boxes to a long and comprehensive visual scene description by the LLM and then evaluates the alignment between the description and the text prompt.
        \item \textbf{LLMScore$^+$} is an improved version of LLMScore. To mitigate excessive hallucination in the long description generated by the LLM, we directly utilize the descriptions from instructblip~\cite{DBLP:journals/corr/abs-2305-06500}. All other modules maintain their original settings as in LLMScore.
        \item \textbf{LLMScore$^{++}$} is an improved version of LLMScore$^+$ with self-verification. LLMs struggle to identify the text-based image evaluation and the image-based text evaluation. To tackle this, we introduce self-verification modules that allow the LLM to verify answer accuracy. First, the LLM serves as an explanation generator, providing compositional error numbers and rationales that must follow a specific format to highlight modality differences: \emph{The image described/did not describe ..., while the text did not mention/did mention ...}. Second, the LLM verifies these errors and rationales via VQA tasks to minimize hallucinations, with questions based on the rationales. Finally, the LLM re-evaluates the image based on the verified rationales.
        
    \end{itemize}

\end{itemize}

\subsection{Human Annotation}

The human rating on images generated by different T2I synthesis models is used to evaluate the quality of automatic metrics. 
%
%
We construct a subset for human rating. The subset is called \emph{sample-100}, randomly sampling 5-6 cases for each category and a total sample of 100 sentence pairs (200 contrastive sentences). We generate one image using SD1.5, SD2.1, SDXL, and IF and collect 800 image-text pairs for human rating by 3 annotators. 

\begin{table}[ht]
\footnotesize
\setlength\tabcolsep{3pt} 
\centering
\begin{tabular}{ccccc|cc}
\hline
\multirow{2}{*}{$h_{point}(\cdot)$} & \multirow{2}{*}{$g_{point}(\cdot)$} & \multirow{2}{*}{\begin{tabular}[c]{@{}l@{}}LLMScore\\Version\end{tabular}}  & \multicolumn{2}{c|}{sample100-$T_0$} & \multicolumn{2}{c}{sample100-$T_1$}\\ \cline{4-7} 
 & & & $\tau (\uparrow) $ & $\rho (\uparrow)$ & $\tau (\uparrow) $ & $\rho (\uparrow)$ \\ \hline
\multirow{4}{*}{$h_{10}(\cdot)$} & \multirow{2}{*}{$g_{100}(\cdot)$} & + & 0.14 & 0.18 & 0.11 & 0.14 \\ \cline{4-7} 
& & ++ & 0.16 & 0.22 & 0.14  & 0.18 \\ \cline{4-7} 
& \multirow{2}{*}{$g_{5}(\cdot)$} & + & 0.20 & 0.25 & 0.19 & 0.21 \\ \cline{4-7} 
& & ++ & 0.24 & \underline{0.30} & \underline{0.27} & \underline{0.32} \\ \hline

\multirow{4}{*}{$h_{5}(\cdot)$} & \multirow{2}{*}{$g_{100}(\cdot)$} & + & 0.17 & 0.20 & 0.14 & 0.16 \\ \cline{4-7} 
& & ++ & 0.23 & 0.28 & 0.22 & 0.27 \\ \cline{4-7} 
& \multirow{2}{*}{$g_{5}(\cdot)$} & + & \underline{0.26} & 0.29 & 0.19 & 0.22 \\ \cline{4-7} 
& & ++ & \textbf{0.35} & \textbf{0.41} & \textbf{0.29} & \textbf{0.33} \\ \hline
\end{tabular}
\caption{Results of 5-point~\cite{DBLP:journals/corr/abs-2307-06350} and 10-point~\cite{DBLP:journals/corr/abs-2305-11116} human rating on LLMScore$^+$ and LLMScore$^{++}$. Images are generated by SD 1.5.
\label{tab:non-commutative}
}
\end{table}

\begin{figure}[ht]
  \centering
   \includegraphics[width=1\linewidth]{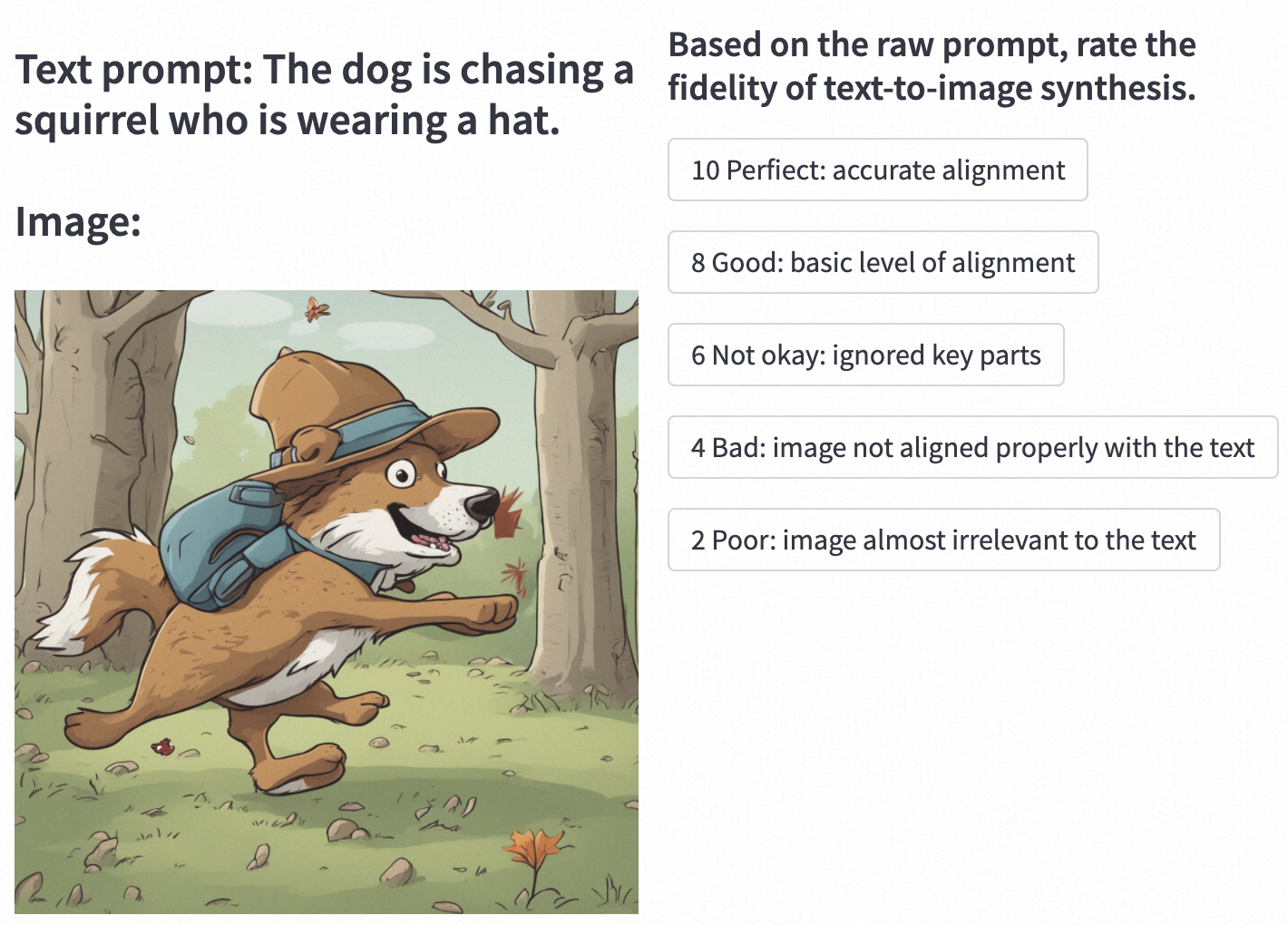}

   \caption{Interface for the human ratings}
   \label{fig:annotation_interface}
\end{figure}

During human annotation, the 10-point scale used for LLMScore\cite{DBLP:journals/corr/abs-2305-11116} was found overly detailed and challenging for annotators. As a result, we adopted a more practical and intuitive 5-point scale, similar to the one used for MiniGPT4-CoT's complex compositions\cite{DBLP:journals/corr/abs-2307-06350}. Our experiments indicate that an overly detailed rating system can significantly influence correlation coefficients, corroborating annotators' intuition, as demonstrated in Table \ref{tab:non-commutative}. However, for consistency with other studies, we use LLMScore's standard 100-point scale for automatic ratings and a 5-point scale for human ratings. Figure \ref{fig:annotation_interface} depicts the human evaluation interface.

\section{Additional Experiment Results}

\begin{figure}[ht]
  \centering
  
   \begin{subfigure}[b]{0.5\linewidth}
        \centering
        \includegraphics[width=\linewidth]{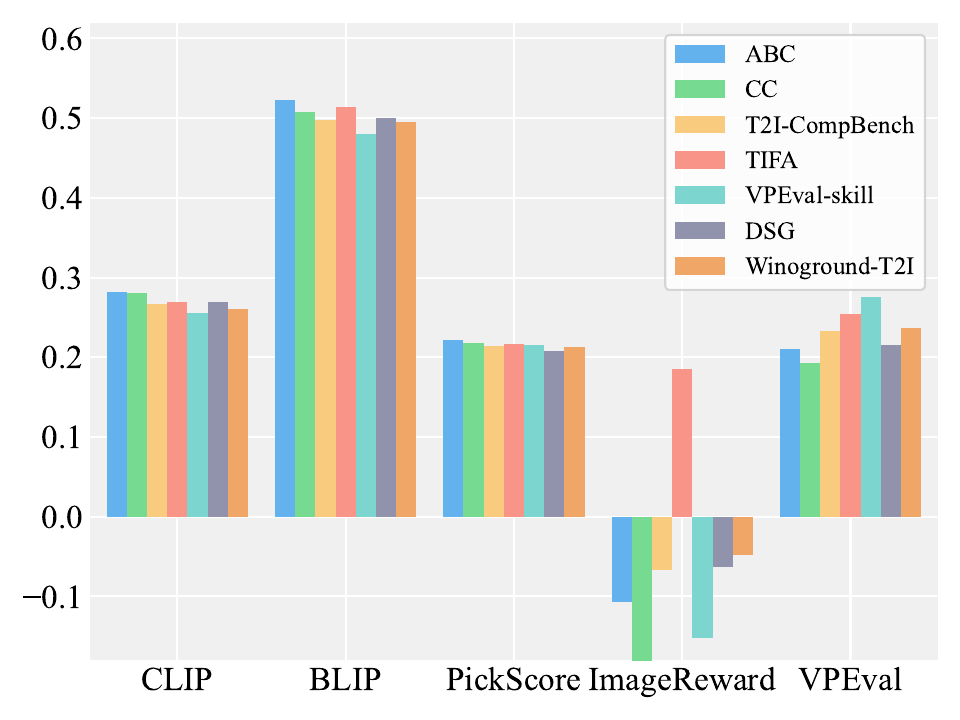}
        \label{fig:benchmark_sd15_result_1}
   \end{subfigure}%
   \begin{subfigure}[b]{0.5\linewidth}
        \centering
        \includegraphics[width=\linewidth]{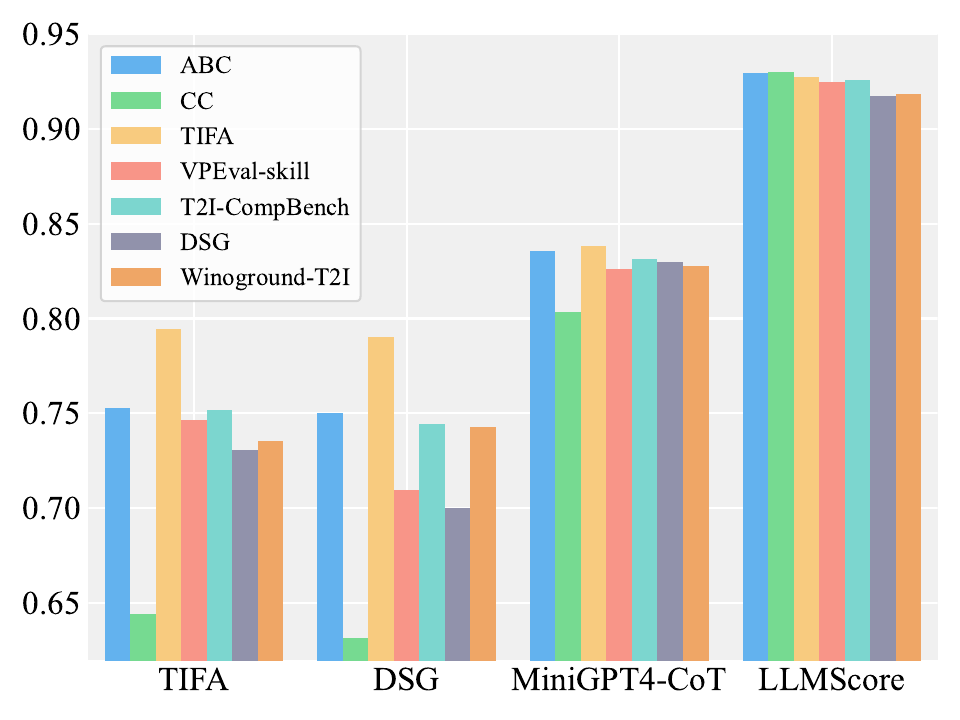}
        \label{fig:benchmark_sd15_result_2}
   \end{subfigure}
   \caption{Evaluation results from 7 benchmarks on SD1.5. The ImageReward score of the CC benchmark is -0.5, where scores less than -0.18 have been omitted.}
   \label{fig:benchmark_sd15_results}
\end{figure}

\begin{figure}[ht]
  \centering
  
   \begin{subfigure}[b]{0.5\linewidth}
        \centering
        \includegraphics[width=\linewidth]{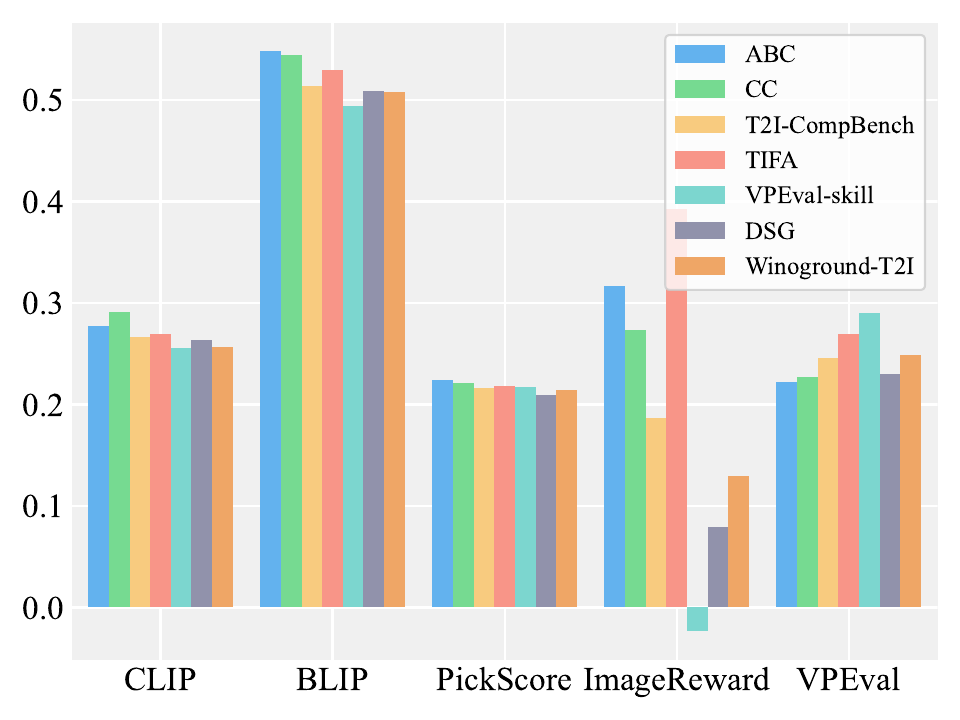}
        \label{fig:benchmark_sd21_result_1}
   \end{subfigure}%
   \begin{subfigure}[b]{0.5\linewidth}
        \centering
        \includegraphics[width=\linewidth]{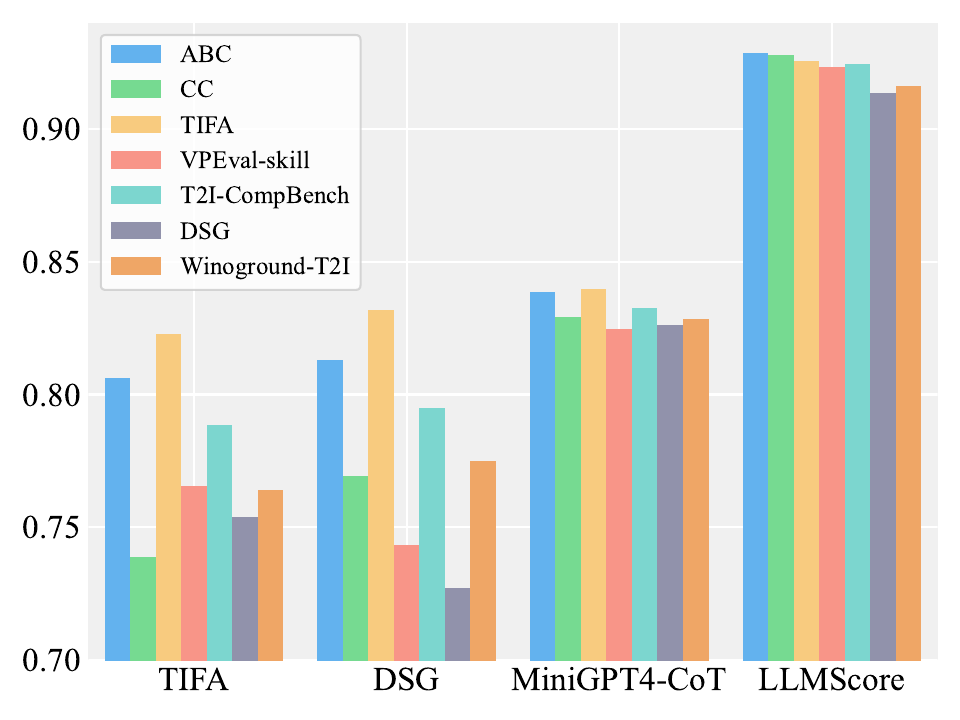}
        \label{fig:benchmark_sd21_result_2}
   \end{subfigure}
   \caption{Evaluation results from 7 benchmarks on SD2.1.}
   \label{fig:benchmark_sd21_results}
\end{figure}

\begin{figure}[ht]
  \centering
  
   \begin{subfigure}[b]{0.5\linewidth}
        \centering
        \includegraphics[width=\linewidth]{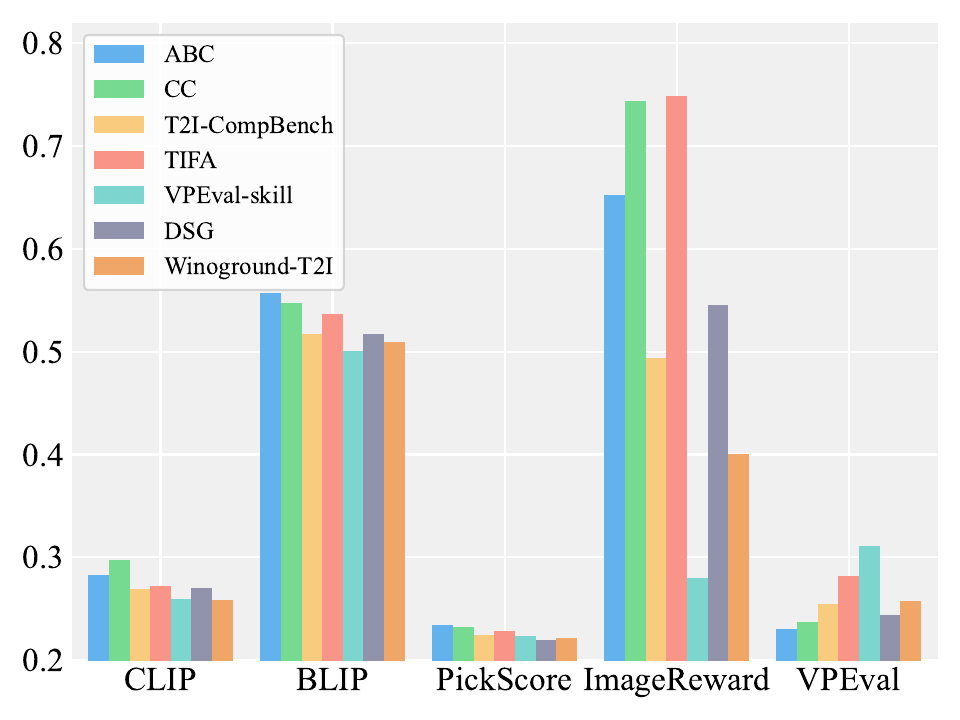}
        \label{fig:benchmark_sdxl_result_1}
   \end{subfigure}%
   \begin{subfigure}[b]{0.5\linewidth}
        \centering
        \includegraphics[width=\linewidth]{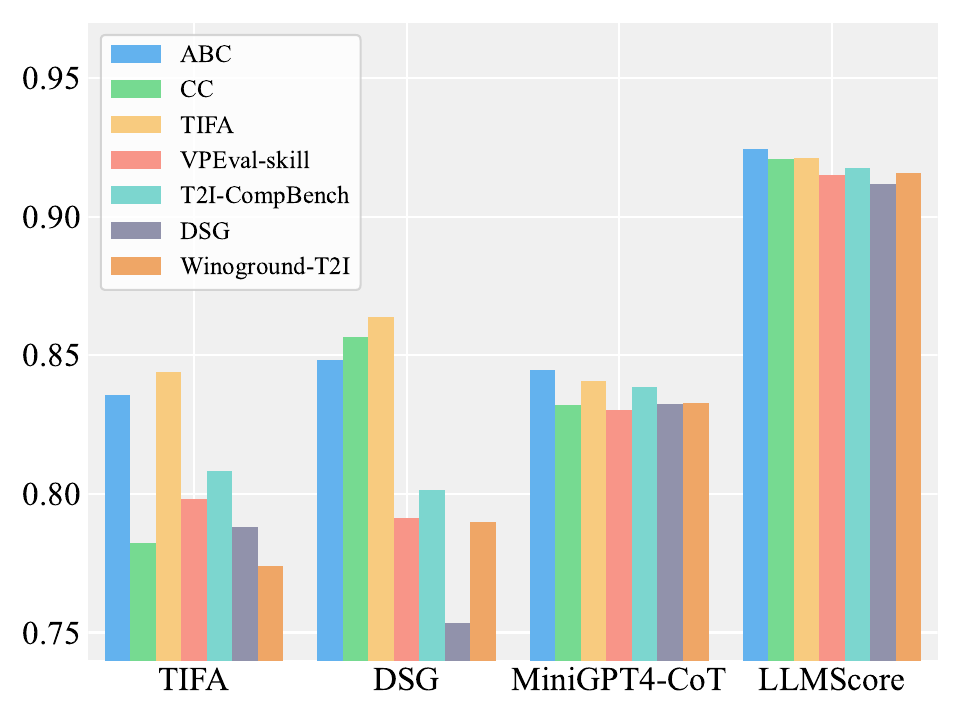}
        \label{fig:benchmark_sdxl_result_2}
   \end{subfigure}
   \caption{Evaluation results from 7 benchmarks on SDXL.}
   \label{fig:benchmark_sdxl_results}
\end{figure}

\begin{figure}[ht]
  \centering
  
   \begin{subfigure}[b]{0.5\linewidth}
        \centering
        \includegraphics[width=\linewidth]{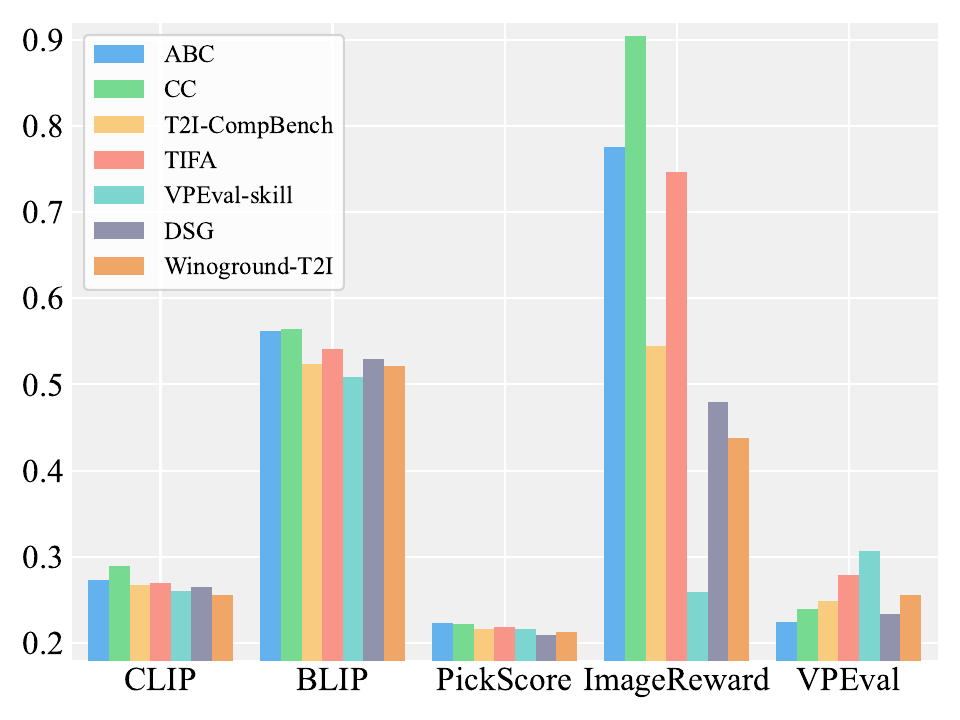}
        \label{fig:benchmark_if_result_1}
   \end{subfigure}%
   \begin{subfigure}[b]{0.5\linewidth}
        \centering
        \includegraphics[width=\linewidth]{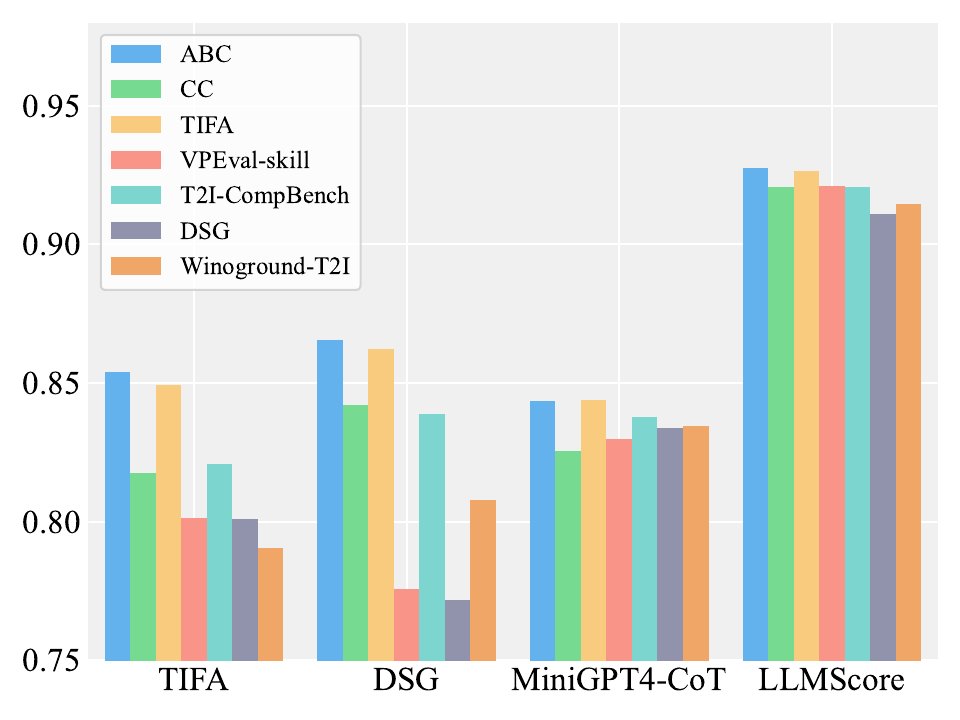}
        \label{fig:benchmark_if_result_2}
   \end{subfigure}
   \caption{Evaluation results from 7 benchmarks on IF.}
   \label{fig:benchmark_if_results}
\end{figure}

\noindent
\textbf{Evaluating Benchmarks}. We compare 7 benchmarks, each evaluated using 9 metrics on SD1.5, SD2.1, SDXL, and IF. The results are shown in Figure \ref{fig:benchmark_sd15_results}, Figure \ref{fig:benchmark_sd21_results}, Figure \ref{fig:benchmark_sdxl_results}, and Figure \ref{fig:benchmark_if_results}. Although the performance of these benchmarks varies across different metrics, it is noticeable that in certain metrics, the scores of VPEval-skill and DSG are lower than ours, particularly in the ImageReward and DSG metrics.

\begin{table}[ht]
\footnotesize
\setlength\tabcolsep{3pt} 
\centering
\begin{tabular}{llllllll}
\hline
\begin{tabular}[c]{@{}l@{}}count/\\length\end{tabular} & ABC   & CC    & \begin{tabular}[c]{@{}l@{}}T2I-\\ Comp\\Bench\end{tabular} & TIFA  & \begin{tabular}[c]{@{}l@{}}VP\\Eval-\\ skill\end{tabular} & DSG   & \begin{tabular}[c]{@{}l@{}}Wino\\ground-\\ T2I\end{tabular} \\
\hline
word  & 12.81 & 7.00 & 9.60 & 10.47 & 6.61 & 17.14 & 12.04 \\
sentence & 61.81 & 28.88 & 49.15 & 52.77 & 29.61 & 92.48 & 62.01 \\                 
\hline
\end{tabular}
\caption{The average word count and sentence length in each benchmark.
\label{tab:sentence_length}
}
\end{table}

\begin{figure*}[ht]
  \centering
   \includegraphics[width=1\linewidth]{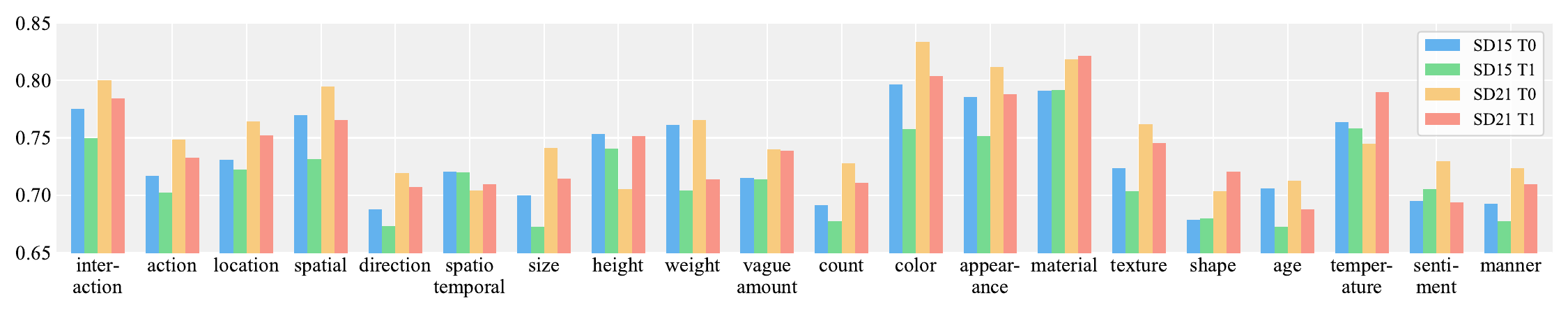}

   \caption{Evaluation results of categories of Winoground-T2I from SD15 and SD21}
   \label{fig:categories_results_additional}
\end{figure*}

In analyzing the data distributions of VPEval-skill and DSG, we notice that DSG-1k's average sentence length is 17 words, 50\% longer than ours, complicating T2I model generation and reducing image scores, as shown in Table \ref{tab:sentence_length}. 
In contrast, VPEval-skill's average is only 6.6 words, about half of ours, with 47\% of prompts being simple single-object, yet scoring low. 
We aim to investigate the reason for the low scores of the VPEval-skill benchmark, composed of simple and short sentences, in contrast to the higher scores achieved by similarly short sentences in CC.

Our first finding suggests that ImageReward may not serve as an ideal evaluation metric for the compositionality of T2I models. First, the performance of ImageReward in object recognition is less satisfied. It is trained on SD1.4-generated realistic-style images, limiting its effectiveness for subsequent models. An analysis of 100 \emph{Object} category cases revealed a 6.5\% accuracy despite 62\% of SDXL-generated images receiving negative ImageReward ratings. Similarly, SD1.5, SD2.1, and IF-generated images had negative ratings of 56\%, 71\%, and 48\%, with accuracy rates of 16.7\%, 18.3\%, and 10.42\%, respectively. Secondly, we found that ImageReward scores for equally aligned images exhibit inconsistency, ranging from approximately -2 to 2. The metric considers not only alignment but also aesthetics.

Our second finding suggests that DSG has proven its validity in long sentences by emphasizing primary sentence elements~\cite{DBLP:journals/corr/abs-2310-18235}. However, in short sentences, each element significantly influences the overall score, and DSG further amplifies the score of these primary elements. If an object is not depicted, all related relationships receive a score of zero, thus reducing the overall score. For instance, all sentences in the \emph{Scale} category of VPEval-skill, which comprise two objects and their relative size, could lose 1/3-1/4 of the score if any part is missing. Therefore, for shorter sentences, the TIFA metric provides a more accurate indication of generation quality due to its ability to lessen the influence of sentence length.

\begin{table}[htbp]
\centering

\begin{subtable}{\columnwidth}
\scriptsize
\centering
\setlength\tabcolsep{3pt} 
\begin{tabular}{ccc|ccc|ccccc}
\hline
\multirow{3}{*}{Metric} & \multicolumn{2}{c|}{$r_{same}$} & \multicolumn{3}{c|}{$p_{same}(\uparrow)$} & \multicolumn{5}{c}{$p_{diff}(\uparrow)$} \\ \cline{2-11}
 & \multirow{2}{*}{$\tau(\uparrow)$} & \multirow{2}{*}{$\rho(\uparrow)$} & $\epsilon=$ & $\epsilon=$ & $\epsilon=$ & $\Omega=$ & $\Omega=$ & $\Omega=$ & $\Omega=$ & $\Omega=$ \\ 
 & & & 0.05 & 0.07 & 0.1 & 0.4 & 0.5 & 0.6 & 0.7 & 0.8 \\ \hline
CLIPScore  & 0.44 & 0.59 & 0.18 & 0.27 & 0.49 & 0.04 & -- & -- & -- & -- \\ 
BLIP-ITM   & 0.40 & 0.54 & 0.27 & 0.37 & 0.53 & 0.05 & 0.07 & 0.12 & -- & -- \\ 
PickScore  & \textbf{0.53} & \textbf{0.70} & 0.31 & 0.41 & \textbf{0.69} & 0.07 & -- & -- & -- & -- \\ 
ImageReward & 0.46 & 0.63 & 0.33 & 0.33 & 
0.39 & 0.10 & 0.13 & 0.18 & 0.28 & 0.43 \\  
\hline
VPEval(fixed) & \underline{0.52} & 0.62 & \textbf{0.49} & \textbf{0.53} & \underline{0.63} & \textbf{0.77} & \textbf{0.80} & \textbf{0.80} & \textbf{1.00} & -- \\
VPEval(unfixed) & \underline{0.52} & \underline{0.64} & \underline{0.39} & \underline{0.49} & 0.49 & 0.21 & 0.32 & 0.57 & -- & -- \\ 
TIFA(fixed) & 0.37 & 0.47 & \underline{0.39} & 0.41 & 0.41 & \underline{0.56} & \underline{0.60} & \underline{0.63} & \underline{0.70} & \underline{0.76} \\
TIFA(unfixed) & 0.32 & 0.42  & 0.37 & 0.39 & 0.41 & 0.32 & 0.33 & 0.37 & 0.49 & 0.57 \\
DSG(fixed) & 0.23 & 0.27 & 0.35 & 0.35 & 0.37 & 0.50 & 0.51 & 0.58 & 0.62 & 0.74 \\
DSG(unfixed) & 0.27 & 0.33  & 0.35 & 0.37 & 0.51 & 0.38 & 0.38 & 0.45 & 0.58 & 0.64 \\
\hline
MiniGPT-CoT   & 0.35 & 0.45 & 0.14 & 0.14 & 0.45 & 0.38 & 0.40 & 0.41 & 0.50 & 0.66 \\ 
LLMScore & 0.22 & 0.27 & 0.16 & 0.16 & 0.20 & 0.34 & 0.43 & 0.43 & 0.51 & \textbf{1.00} \\ 
LLMScore$^+$ & 0.29 & 0.35 & 0.29 & 0.29 & 0.29 & -- & -- & -- & -- & -- \\
LLMScore$^{++}$ & 0.21 & 0.26 & 0.24 & 0.24 & 0.24 & 0.25 & 0.24 & 0.44 & -- & -- \\ 
\hline
\end{tabular}
\caption{SD2.1.
\label{tab:consistency_sd21}
}
\end{subtable}

\vspace{1em} 

\begin{subtable}{\columnwidth}
\scriptsize
\centering
\setlength\tabcolsep{3pt} 
\begin{tabular}{ccc|ccc|ccccc}
\hline
\multirow{3}{*}{Metric} & \multicolumn{2}{c|}{$r_{same}(\uparrow)$} & \multicolumn{3}{c|}{$p_{same}(\uparrow)$} & \multicolumn{5}{c}{$p_{diff}(\uparrow)$} \\ \cline{2-11}
 & \multirow{2}{*}{$\tau(\uparrow)$} & \multirow{2}{*}{$\rho(\uparrow)$} & $\epsilon=$ & $\epsilon=$ & $\epsilon=$ & $\Omega=$ & $\Omega=$ & $\Omega=$ & $\Omega=$ & $\Omega=$ \\ 
 & & & 0.05 & 0.07 & 0.1 & 0.4 & 0.5 & 0.6 & 0.7 & 0.8 \\ \hline
CLIPScore  & 0.24 & 0.34 & 0.14 & 0.21 & 0.33 & 0.10 & 0.08 & -- & -- & -- \\ 
BLIP-ITM   & 0.29 & 0.44 & 0.14 & 0.21 & 0.33 & 0.10 & 0.09 & 0.14 & -- & -- \\ 
PickScore  & \textbf{0.57} & \textbf{0.77} & 0.19 & 0.35 & 0.42 & 0.10 & -- & -- & -- & -- \\ 
ImageReward & 0.34 & 0.48 & 0.21 & 0.35 & 
0.37 & 0.14 & 0.17 & 0.21 & 0.32 & 0.33 \\  
\hline
VPEval(fixed) & 0.45 & 0.58 & \underline{0.40} & 0.42 & 0.47 & \textbf{0.85} & \textbf{0.85} & \textbf{0.92} & \textbf{1.00} & \textbf{1.00} \\
VPEval(unfixed) & \underline{0.46} & \underline{0.60} & 0.35 & \textbf{0.47} & \textbf{0.56} & 0.29 & 0.35 & 0.57 & 0.60 & -- \\ 
TIFA(fixed) & 0.25 & 0.33 & 0.21 & 0.23 & 0.37 & \underline{0.61} & \underline{0.63} & \underline{0.69} & 0.71 & \underline{0.94} \\
TIFA(unfixed) & 0.24 & 0.33 & 0.21 & 0.23 & 0.33 & 0.40 & 0.42 & 0.49 & 0.67 & 0.89 \\
DSG(fixed) & 0.44 & 0.53 & \textbf{0.47} & \textbf{0.47} & 0.47 & 0.60 & 0.60 & 0.63 & \underline{0.74} & 0.81 \\
DSG(unfixed) & 0.36 & 0.48 & \underline{0.40} & \underline{0.44} & \underline{0.49} & 0.35 & 0.37 & 0.44 & 0.51 & 0.67 \\
\hline
MiniGPT-CoT   & 0.04 & 0.06 & 0.21 & \textbf{0.47} & 0.47 & 0.47 & 0.47 & 0.51 & 0.60 & \textbf{1.00} \\ 
LLMScore & 0.25 & 0.32 & 0.19 & 0.21 & 0.23 & 0.27 & 0.28 & 0.37 & 0.45 & 0.71 \\ 
LLMScore$^+$ & 0.12 & 0.15 & 0.35 & 0.35 & 0.35 & -- & -- & -- & -- & -- \\
LLMScore$^{++}$ & 0.42 & 0.47 & 0.42 & 0.42 & 0.42 & 0.29 & 0.30 & 0.25 & 0.43 & -- \\ 
\hline
\end{tabular}
\caption{IF.
\label{tab:consistency_if}
}
\end{subtable}

\caption{Results of intra-pair comparison on sample-100. The images are generated by SD2.1 and IF.}
\label{tab:consistency_additional}

\end{table}

\begin{table}[ht]
\scriptsize
\centering
\setlength\tabcolsep{3pt} 
\begin{tabular}{ccc|cc|cc|cc}
\hline
\multirow{3}{*}{Metric} & \multicolumn{4}{c|}{SD2.1} & \multicolumn{4}{c}{IF} \\ \cline{2-9}
 & \multicolumn{2}{c|}{$T_0$} & \multicolumn{2}{c|}{$T_1$} & \multicolumn{2}{c|}{$T_0$} & \multicolumn{2}{c}{$T_1$} \\ \cline{2-9}
 & $\bar \tau(\uparrow)$ & $\bar \rho(\uparrow)$ & $\bar \tau(\uparrow)$ & 
 $\bar \rho(\uparrow)$ & $\bar \tau(\uparrow)$ & $\bar \rho(\uparrow)$ & $\bar \tau(\uparrow)$ & $\bar \rho(\uparrow)$ \\  
\hline
CLIPScore  & \textbf{1.00} & \textbf{1.00} & \textbf{1.00} & \textbf{1.00} & \textbf{1.00} & \textbf{1.00} & \textbf{1.00} & \textbf{1.00} \\ 
BLIP-ITM   & \textbf{1.00} & \textbf{1.00} & \textbf{1.00} & \textbf{1.00} & \textbf{1.00} & \textbf{1.00} & \textbf{1.00} & \textbf{1.00} \\ 
PickScore  & \textbf{1.00} & \textbf{1.00} & \textbf{1.00} & \textbf{1.00} & \textbf{1.00} & \textbf{1.00} & \textbf{1.00} & \textbf{1.00} \\ 
ImageReward & \textbf{1.00} & \textbf{1.00} & \textbf{1.00} & \textbf{1.00} & \textbf{1.00} & \textbf{1.00} & \textbf{1.00} & \textbf{1.00} \\  
\hline
VPEval(fixed) & \textbf{1.00} & \textbf{1.00} & \textbf{1.00} & \textbf{1.00} & \textbf{1.00} & \textbf{1.00} & \underline{0.99} & \textbf{1.00} \\
VPEval(unfixed) & 0.92 & 0.96 & 0.92 & 0.95 & 0.88 & 0.93 & 0.91 & 0.95 \\
TIFA(fixed) & \textbf{1.00} & \textbf{1.00} & \textbf{1.00} & \textbf{1.00} & \textbf{1.00} & \textbf{1.00} & \textbf{1.00} & \textbf{1.00} \\
TIFA(unfixed) & \underline{0.97} & \underline{0.99} & 0.97 & \underline{0.99} & \underline{0.95} & \underline{0.97} & 0.94 & 0.96 \\ 
DSG(fixed) & \textbf{1.00} & \textbf{1.00} & \underline{0.99} & \textbf{1.00} & \textbf{1.00} & \textbf{1.00} & \underline{0.99} & \underline{0.99} \\
DSG(unfixed) & 0.84 & 0.93 & 0.85 & 0.91 & 0.74 & 0.83 & 0.75 & 0.83 \\
\hline
MiniGPT-CoT   & 0.08 & 0.09 & 0.11 & 0.13 & 0.09 & 0.10 & 0.10 & 0.11 \\ 
LLMScore & 0.01 & 0.01 & 0.17 & 0.19 & 0.06 & 0.07 & 0.13 & 0.14 \\ 
LLMScore$^+$ & 0.26 & 0.27 & 0.38 & 0.40 & 0.28 & 0.29 & 0.20 & 0.20 \\
LLMScore$^{++}$ & 0.26 & 0.28 & 0.27 & 0.30 & 0.33 & 0.36 & 0.23 & 0.25 \\ 
\hline
\end{tabular}
\caption{Results of stability on sample-100. The images are generated by SD2.1 and IF.
\label{tab:stability_additional}
}
\end{table}

\noindent
\textbf{Evaluating Metrics}. We evaluate the reliability of various metrics on SD1.5, SD2.1, SDXL and IF from four different perspectives. The results of the inter-pair comparison are shown in Table \ref{tab:correlation_1}. The results of the intra-pair comparison are shown in Table \ref{tab:consistency} and Table \ref{tab:consistency_additional}. The results of stability and efficiency are shown in Table \ref{tab:stability} and Table \ref{tab:consistency_additional}.

The metric's performance is consistent across SD1.5, SD2.1, SDXL and IF.
In the efficiency statistics, the time is based on the evaluation results of images generated by SD1.5. 
Despite a slight increase in image loading time due to higher resolution in methods like SD2.1, SDXL, and IF, the magnitude of time and the trend align with SD1.5 results. Consequently, we did not statistic the time cost separately for the other three models.

\noindent
\textbf{Evaluating T2I models across Categories}. We evaluate the performance of SD1.5, SD2.1, SDXL and IF across categories, as shown in Figure \ref{fig:categories_results} and Figure \ref{fig:categories_results_additional}. 
From 1.5 to IF, the models' capabilities in each category have gradually improved. However, regardless of the model, \emph{Interaction}, \emph{Spatial}, \emph{Color}, \emph{Appearance}, and \emph{Material} remain the categories of strength, with no change in the differences among these categories.


\end{document}